\pdfoutput=1

\documentclass[11pt]{article}

\usepackage[final]{acl}

\usepackage{times}
\usepackage{latexsym}

\usepackage[T1]{fontenc}

\usepackage[utf8]{inputenc}

\usepackage{microtype}

\usepackage{inconsolata}

\usepackage{graphicx}
\usepackage{amsfonts}
\usepackage{amsmath} 
\usepackage{tabularx, booktabs}
\usepackage{algorithm}
\usepackage{algpseudocode}
\usepackage{colortbl}
\usepackage{bm}
\usepackage{multirow}
\usepackage[most]{tcolorbox}
\usepackage{pifont}

\usepackage{subfigure}
\usepackage{booktabs} 
\usepackage{enumitem}
\usepackage{makecell} 

\tcbuselibrary{listingsutf8}
\usepackage{listings}
\usepackage{xcolor}
\usepackage{adjustbox}

\definecolor{promptgray}{gray}{0.95}
\lstdefinelanguage{Prompt}{
  basicstyle=\ttfamily\footnotesize,
  backgroundcolor=\color{promptgray},
  breaklines=true,
  frame=single,
  showstringspaces=false,
}

\definecolor{lightgray}{gray}{0.95}
\title{Dropping Experts, Recombining Neurons: \\ Retraining-Free Pruning for Sparse Mixture-of-Experts LLMs}

\author{
 \textbf{Yixiao Zhou\textsuperscript{1,2}},
 \textbf{Ziyu Zhao\textsuperscript{1,2}},
 \textbf{Dongzhou Cheng\textsuperscript{2,3}},
 \textbf{Zhiliang Wu\textsuperscript{1}},
\\
 \textbf{Jie Gui\textsuperscript{3}},
 \textbf{Yi Yang\textsuperscript{1}},
 \textbf{Fei Wu\textsuperscript{1,4}},
 \textbf{Yu Cheng\textsuperscript{2,5}}\thanks{Corresponding authors.},
 \textbf{Hehe Fan\textsuperscript{1}}\footnotemark[1]
\\
 \textsuperscript{1}Zhejiang University,
 \textsuperscript{2}Shanghai Innovation Institute,
 \textsuperscript{3}Southeast University,
 \\
 \textsuperscript{4}Shanghai AI Laboratory,
 \textsuperscript{5}The Chinese University of Hong Kong
\\
\texttt{\{12421181,ziyuzhao.cs,wu\_zhiliang,yangyics,wufei,hehefan\}@zju.edu.cn},
\\
\texttt{\{230249457,guijie\}@seu.edu.cn}, \texttt{\{chengyu\}@cse.cuhk.edu.hk}
}

\begin{document}
\maketitle
\begin{abstract}
Sparse Mixture-of-Experts (SMoE) architectures are widely used in large language models (LLMs) due to their computational efficiency. However, though only a few experts are activated for each token, SMoE still requires loading all expert parameters, leading to high memory usage and challenges in deployment. Previous work has tried to reduce the overhead by pruning and merging experts, but primarily focused on expert-level operations, leaving neuron-level structure underexplored. We propose \textbf{DERN} (\textbf{D}ropping \textbf{E}xperts, \textbf{R}ecombining \textbf{N}eurons), a task-agnostic and retraining-free framework for expert pruning and reconstruction. We observe that experts are often misaligned and contain semantic conflicts at the neuron level, which poses challenges for direct merging. To solve this, DERN works in three steps: it first prunes redundant experts using router statistics; then it decomposes them into neuron-level expert segments, assigning each segment to its most compatible retained expert; and finally, it merges segments within each retained expert to build a compact representation. Experiments on Mixtral, Qwen, and DeepSeek SMoE models show that DERN achieves over a 5\% performance gains than previous methods on commonsense reasoning and MMLU benchmarks under 50\% expert sparsity, without extra training. It also greatly reduces the number of experts and memory usage, making SMoE LLMs easier to deploy in practice.

\end{abstract}

\section{Introduction}
Large Language Models (LLMs) have become the foundation models of modern NLP~\cite{brown2020language}, with their application rapidly expanding into diverse domains such as multimodal learning~\cite{zhang2025prompt, li2024topa, li2024improving, li2024cat}, scientific research~\cite{wang2025protchatgpt, hu2025osda}, and advanced reasoning~\cite{dong2025enhancing}. This widespread use, however, makes their ever-growing scale a serious challenge for efficient deployment.
Sparse Mixture-of-Experts (SMoE) architectures alleviate inference cost by activating only a subset of experts per token~\cite{shazeer2017outrageously, fedus2022switch}, matching or surpassing dense models in performance. However, their total parameter footprint remains massive. For instance, DeepSeek-V3~\cite{liu2024deepseek} activates only 37B parameters per pass, yet requires storing 671B in memory. Reducing this overhead without compromising performance remains an open challenge.

\begin{table}[tb]
\caption{Comparison of our method with other expert pruning methods. E-Merge means expert-level merge, while \textbf{N-Merge} means neuron-level merge.}
\vspace{-3mm}
\label{tab:methods}
\footnotesize
\centering
\begin{adjustbox}{width=\linewidth}
\begin{tabular}{l@{}ccc}
\toprule
\textbf{Method} & \textbf{Task-Agnostic} & \textbf{Train-Free} & \textbf{Strategy} \\
\midrule
TSEP~\cite{chen2022task}             & \ding{55} & \ding{55} & Prune \\
MoE-$I^2$~\cite{yang2024moe}         & \ding{51} & \ding{55} & Prune \\
NAEE~\cite{lu2024not}                & \ding{51} & \ding{51} & Prune \\
HC-SMoE~\cite{chen2024retraining}    & \ding{51} & \ding{51} & E-Merge \\
MC-SMoE~\cite{li2023merge}           & \ding{51} & \ding{55} & E-Merge \\
\textbf{DERN (Ours)}                 & \ding{51} & \ding{51} & \textbf{N-Merge} \\
\bottomrule
\end{tabular}
\end{adjustbox}
\vspace{-4mm}
\end{table}

Recent work has shown that MoE layers exhibit significant redundancy: experts contribute unequally to downstream predictions~\cite{chi2022representation}, and many share high parameter similarity~\cite{lo2024closer}, indicating overlapping functional capacity and structural duplication. These findings motivate expert-level sparsification as a promising direction. Early approaches, such as ~\cite{chen2022task} prune experts progressively during fine-tuning on a specific task, but suffer from high training overhead. Subsequent approaches shifted toward retraining-free and task-agnostic pruning. For example, ~\cite{lu2024not} searches for optimal expert subsets using the output discrepancy, while ~\cite{he2024demystifying} leverages routing statistics for large-scale pruning. However, the abrupt removal of expert parameters can significantly impair the model's capabilities.

\begin{figure*}[htb!]
    \centering  \includegraphics[width=1\linewidth]{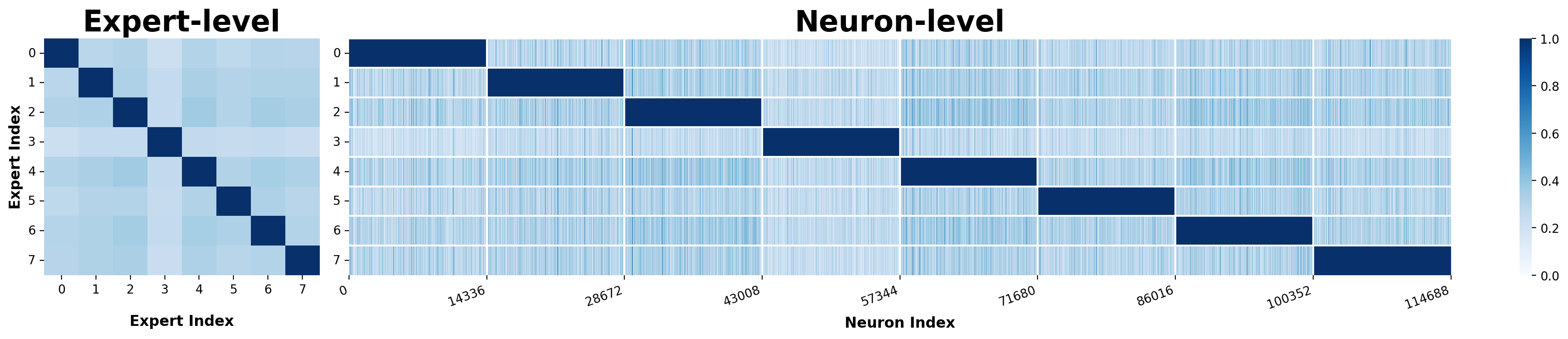}
    \caption{Expert-level and neuron-level cosine similarity in layer 15 of Mixtral-8×7B-Instruct. The left plot shows expert-to-expert similarity; the right plot depicts how strongly a neuron aligns with the most similar neuron in each expert. Despite some global expert alignment (left), clear inconsistencies remain at the neuron level (right).}
    \label{fig:sim-neuron}
\vspace{-4mm}
\end{figure*}

To mitigate such loss, a parallel line of work explores expert merging as a post-pruning remedy. Representative methods~\cite{li2023merge, chen2024retraining, liu2024efficient} group or cluster experts followed by weighted averaging. However, these approaches merge at the whole-expert level, often overlooking that neuron arrangements across experts are inherently unaligned~\cite{ainsworth2022git}. Even with alignment mechanisms~\cite{li2023merge}, similar experts may encode distinct or even conflicting internal representations, due to divergent learned semantics~\cite{zhao2024merging, zhao2025each, ruan2025task}. Such structural and semantic mismatches limit the effectiveness of direct expert fusion methods and can introduce nontrivial degradation in merged models.

To probe this issue, we visualize the similarity at the expert level and the neuron level for Mixtral~\cite{jiang2024mixtral} in Fig.~\ref{fig:sim-neuron}. The results reveal that, while some experts exhibit global alignment in their structural patterns, significant inconsistencies remain at the neuron level. This indicates that expert merging is feasible in principle, but naive whole-expert averaging may fail to preserve neuron-level consistency. Is it possible to enable expert merge through neuron-level, structure-aware recombination of transferable components?

In this paper, we introduce \textbf{DERN} (\textbf{D}ropping \textbf{E}xperts, \textbf{R}ecombining \textbf{N}eurons), a new paradigm for expert pruning and reconstruction, grounded in segment-based modularity and recomposition. At the core of DERN is the notion of an \textbf{\emph{Expert Segment}}: a minimal functional triplet composed of corresponding rows from the gate and up-projection matrices and a column from the down-projection matrix, as detailed in Sec.~\ref{sec:segment-decomposition}. Then each expert is viewed as a collection of such segments.

\begin{figure*}[htb]
    \centering
    \includegraphics[width=1.0\linewidth]{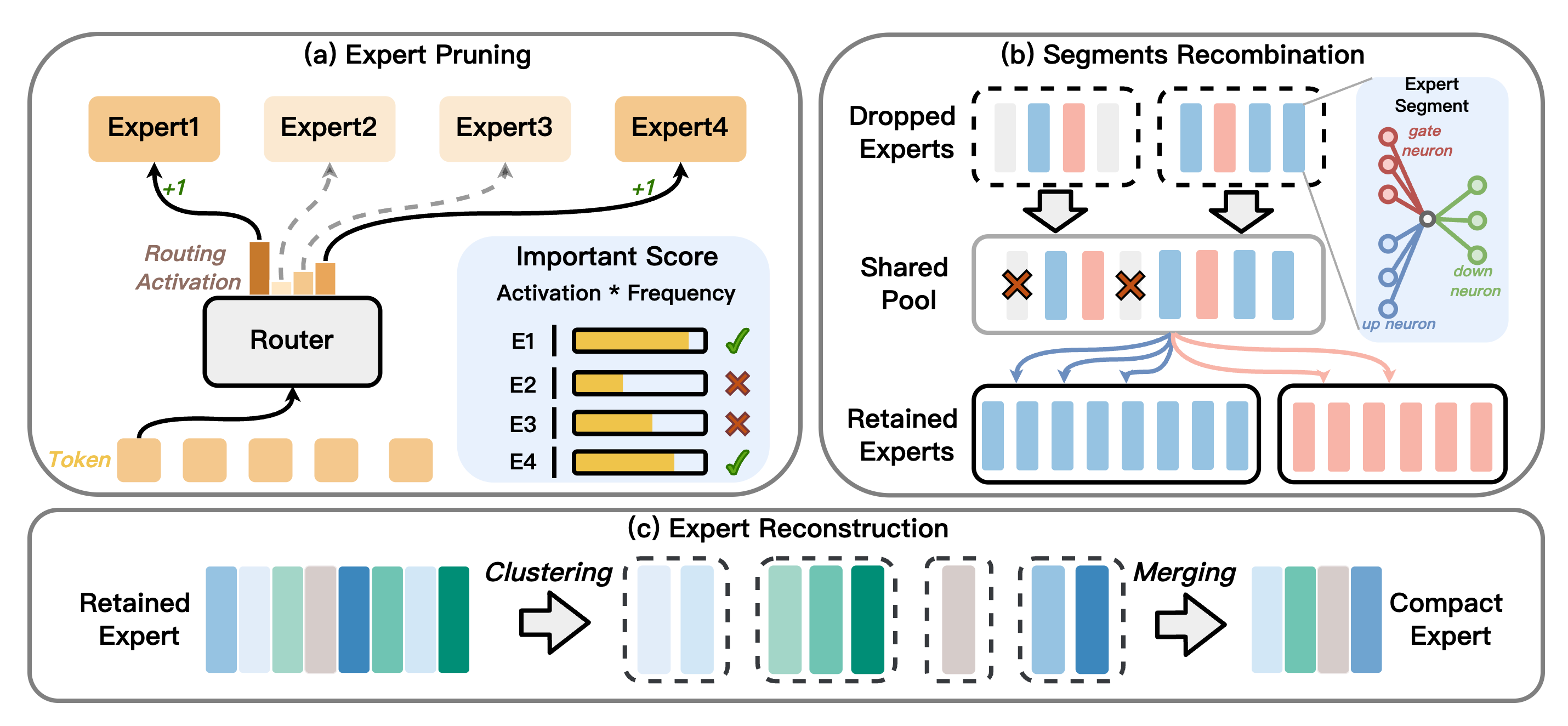}
    \caption{Overview of the \textbf{DERN} framework. (a) Redundant experts are pruned based on routing activation statistics. (b) Dropped experts are decomposed into neuron-level segments, each represented by a triplet of gate, up, and down projection vectors. These segments are then reassigned to retained experts based on local structural similarity. (c) Within each retained expert, reassigned and original segments are clustered via spherical weighted $k$-means to form a compact, performance-preserving expert.}
    \label{fig:dern}
\vspace{-4mm}
\end{figure*}

As illustrated in Fig.~\ref{fig:dern}, DERN proceeds in three stages:  
(1) It identifies and prunes redundant experts based on routing activation statistics;  
(2) It decomposes pruned experts into segments, pools them, and reassigns each segment to the most compatible retained expert based on local structural similarity;  
(3) It applies spherical weighted $k$-Means clustering~\cite{dhillon2001concept} to merge segments within each retained expert, using cluster centroids to reconstruct more compact yet expressive experts with significantly fewer parameters.

Unlike previous whole-expert weighted averaging~\cite{li2023merge, chen2024retraining, liu2024efficient}, DERN reframes pruning as a segment-based decomposition and recombination problem. This modular view enables flexible, fine-grained knowledge transfer across experts, breaking the rigidity of fixed expert partitions, and enhancing the overall expressiveness of pruned models. Our main contributions are summarized as follows:
\begin{itemize}
  \item We reformulate expert merging as a segment-based decomposition and recombination problem, enabling structure-aware, neuron-level operation that surpasses conventional coarse-grained averaging.
  \item We propose \textbf{DERN}, a unified, task-agnostic, and retraining-free pruning framework that restructures experts through a multi-stage pipeline of identification, segments decomposition, adaptive recombination, and cluster-based reconstruction. 
  \item We conduct extensive experiments on Mixtral, Qwen, and DeepSeek SMoE models across multiple commonsense reasoning datasets and the comprehensive MMLU benchmark, demonstrating that DERN achieves over a 5\% performance gains than previous methods under 50\% expert sparsity.
\end{itemize}

\section{Related Work}
\subsection{Sparse Mixture-of-Experts}
Sparse Mixture-of-Experts architectures have emerged as a key paradigm for scaling large language models efficiently, offering reduced inference cost by activating only a small subset of sub-networks (experts) per token~\cite{shazeer2017outrageously, fedus2022switch}. In SMoEs, the dense feedforward layers in Transformers are replaced with multiple parallel experts, typically feedforward networks (FFNs), governed by a trainable gating mechanism that dynamically selects a small number of experts for each token. This enables a substantial increase in total model capacity while maintaining inference cost proportional to the number of active experts. SMoEs have been adopted in many recent LLMs, including Mixtral~\cite{jiang2024mixtral}, DeepSeek-MoE~\cite{dai2024deepseekmoe, liu2024deepseek}, and Qwen-MoE~\cite{yang2024qwen2technicalreport}.

\subsection{Expert Pruning and Merging}

Expert pruning and merging, as a SMoE compression technique, have gained growing interest. Expert pruning is motivated by observations of expert redundancy and contribution imbalance~\cite{chi2022representation, lo2024closer}. Early methods such as ~\cite{chen2022task} involve fine-tuning and incur high training cost. More recent work explores retraining-free, task-agnostic approaches based on reconstruction loss~\cite{lu2024not}, routing statistics~\cite{he2024demystifying}, or changes in router norms~\cite{chowdhury2024provably}. ~\cite{yang2024moe} further considers internal sparsity within experts. However, pruning entire experts can result in irreversible loss of important representations.

Expert merging, a technique within model merging~\cite{yang2025mix}, retains the knowledge of pruned experts to mitigate performance loss. Representative methods typically perform expert grouping or clustering followed by weighted expert averaging. For example, ~\cite{chen2024retraining} uses hierarchical clustering based on expert output similarity; ~\cite{liu2024efficient} employs evolutionary strategies to iteratively merge similar experts; and ~\cite{li2023merge} identifies dominant experts based on routing patterns to guide grouping and merging. However, most existing merging strategies operate at the expert level, overlooking neuron-level misalignment and potential semantic conflicts. This motivates finer-grained recomposition strategies, which we address with our proposed DERN framework. Tab.~\ref{tab:methods} summarizes the differences between our method and existing approaches.

Other SMoE compression approaches include weight pruning within experts~\cite{xie2024moe, liang2025seap}, layer compression~\cite{cao2024condense, lu2024reassessing, li2024sglp}, low-rank decomposition~\cite{gu2025delta}, and quantization~\cite{huang2024mc}, alongside general pruning methods~\cite{ma2023llm, lu2024generic}.

\section{Proposed Method}
Our DERN framework performs expert pruning and merge through a three-stage process, as illustrated in Fig.~\ref{fig:dern}. First, we identify and retain the top-$k$ experts based on routing activations (Sec.~\ref{sec:pruning}). Next, pruned experts are decomposed into neuron-level segments and reassigned to retained experts (Sec.~\ref{sec:recombination}). Finally, each expert merges its segments through clustering to reconstruct a compact structure (Sec.~\ref{sec:clustering}). The overall procedure is summarized in Alg.~\ref{alg:dern}.

\subsection{Expert Pruning via Routing Behaviors}
\label{sec:pruning}
We identify redundant experts in SMoE layers based on routing behaviors observed over a calibration set, and select them as pruning candidates.

Let an SMoE layer consist of $N$ experts $\mathcal{E} = \{E_1, \dots, E_N\}$ and a routing network \(G: \mathbb{R}^d \to \mathbb{R}^N\), which outputs soft routing weights \(G(x) = \bigl(G_1(x), \dots, G_N(x)\bigr)\) for each token feature \(x\). Given a calibration set \(\mathcal{C} = \{x_m\}_{m=1}^M\), let \(\mathcal{A}(x_m) \subseteq \{1, \dots, N\}\) denote the index set of the top-$k$ experts selected for \(x_m\) by the router.

We define the \textbf{\textit{importance score}} for  $E_i$ as:
\begin{equation}
\label{eq:importance}
S_i = \mathbb{E}_{x_m \sim \mathcal{C}} \left[
  \frac{\mathbb{I}[i \in \mathcal{A}(x_m)] \cdot G_i(x_m)}
       {\sum_{j \in \mathcal{A}(x_m)} G_j(x_m)}
\right],
\end{equation}
where $\mathbb{I}[\cdot]$ is the indicator function. This score reflects both the frequency and the strength of activation, effectively measuring each expert’s utility under realistic input distributions.

After computing $\{S_i\}_{i=1}^N$, we identify the top-$k$ experts with the highest importance scores as retained experts, while designating the remaining $N-k$ as pruning candidates.

\subsection{Expert Decomposition and Segment Recombination}
\label{sec:recombination}

We reformulate the pruning task as a modular decomposition and recombination problem. Instead of discarding entire experts, we decompose them into smaller functional units called segments, which can be selectively reassigned to compatible retained experts. Segments from pruned experts are collected into a shared pool and reallocated based on local structural similarity. This approach facilitates neuron-level knowledge transfer while preserving the functional coherence of each expert.

\paragraph{Expert-to-Segments Decomposition.}
\label{sec:segment-decomposition}
We begin by formalizing the decomposition of a standard SMoE expert. The current mainstream LLMs adopt the Gated Linear Unit (GLU) architecture for MLP layers, where each expert is represented by three weight matrices: \( W_g \in \mathbb{R}^{h \times d} \), \( W_u \in \mathbb{R}^{h \times d} \), and \( W_d \in \mathbb{R}^{d \times h} \), where \( d \) is the input dimension and \( h \) the intermediate hidden size. Given an input token \( x \in \mathbb{R}^d \), the forward computation is expressed as:
\begin{equation}
\label{eq:forward}
f(x) = W_d\left(\sigma(W_g x) \odot (W_u x)\right),
\end{equation}
which can be rewritten as a sum of independent contributions from each hidden dimension:
\begin{equation}
\label{eq:sum}
f(x) = \sum_{i=1}^h w_{d,i} \cdot \left[\sigma(w_{g,i}^\top x) \cdot (w_{u,i}^\top x)\right],
\end{equation}
where \(\sigma(\cdot)\) is the activation function, \( w_{g,i}^\top \) and \( w_{u,i}^\top \) represent the \( i \)-th rows of \( W_g \) and \( W_u \) respectively, and \( w_{d,i} \) represents the \( i \)-th column of \( W_d \).

As shown in Eq.~\eqref{eq:sum}, the expert output is a sum of functionally independent, low-rank transformations. This decomposition motivates our definition of an \textbf{\emph{Expert Segment}}: a minimal self-contained unit governed by the parameter triplet \( (w_{g,i}, w_{u,i}, w_{d,i}) \). Specifically, the \( i \)-th segment is:
\begin{equation}
\label{eq:segment}
\mathrm{seg}_i = (w_{g,i}, w_{u,i}, w_{d,i}).
\end{equation}

Segments exhibit three key properties that make them well-suited for recombination. First, they are functionally independent, each contributing to a single output dimension. Second, they maintain gradient locality, with gradients depending solely on their own parameters. Third, their structural regularity allows each triplet to be flattened into a fixed-dimensional vector, enabling efficient similarity comparison in parameter space. Leveraging these properties, we decompose both retained and pruned experts into segments, aggregating those from pruned experts into a global segment pool \( \mathcal{P} \) for reassignment.

\paragraph{Segment Recombination via local structural similarity.}

To reassign segments from \( \mathcal{P} \) to appropriate retained experts, we use a similarity-based matching scheme. Let \( \mathrm{seg}_i \in \mathcal{P} \) be a candidate segment and \( E_r \) a retained expert with segment set \( \mathcal{S}_r \). Each segment is vectorized as:
\begin{equation}
\label{eq:vectorized-seg}
v(\mathrm{seg}_i) = 
\begin{bmatrix}
w_{g,i} \\
w_{u,i} \\
w_{d,i}
\end{bmatrix}
\in \mathbb{R}^{3d},
\end{equation}
where each component corresponds to a row or column from the expert's gate, up, and down projection matrices, respectively.

The local structural similarity between \( \mathrm{seg}_i \) and \( E_r \) is defined as the maximum cosine similarity with any segment already in \( E_r \):
\begin{equation}
\label{eq:similarity}
\mathrm{sim}(\mathrm{seg}_i, E_r) = \max_{s' \in \mathcal{S}_r} \frac{ \langle v(\mathrm{seg}_i), v(s') \rangle }{ \| v(\mathrm{seg}_i) \| \| v(s') \| }.
\end{equation}

For each segment $\mathrm{seg}_i \in \mathcal{P}$, we first identify the most similar retained expert $E_r^* = \arg\max_{E_r \in \mathcal{E}_r} \mathrm{sim}(\mathrm{seg}_i, E_r)$. The segment is then reassigned to this expert only if the similarity score exceeds the threshold $\alpha$:
\begin{equation}
\label{eq:assignment}
E_r^* \leftarrow E_r^* \cup \{\mathrm{seg}_i\} \quad \text{if} \quad \mathrm{sim}(\mathrm{seg}_i, E_r^*) > \alpha.
\end{equation}

This local structural similarity criterion serves as a soft gating mechanism, filtering out structurally incompatible segments while maintaining coherent feature composition within each expert.

To preserve routing semantics, we softly transfer the routing contribution from \( E_o \) to \( E_r \), scaled by the number of segments in \( E_o \):
\begin{equation}
\label{eq:gate_update}
\mathrm{G}_{E_r} \leftarrow \mathrm{G}_{E_r} + \frac{1}{n} \cdot \mathrm{G}_{E_o},
\end{equation}
where \( n = |\mathcal{S}_{E_o}| \) is the number of original segments in expert \( E_o \) (\textit{i.e.}, its intermediate size).

\renewcommand{\algorithmicrequire}{\textbf{Input:}}
\renewcommand{\algorithmicensure}{\textbf{Output:}}

\begin{algorithm}[tb]
\caption{The Overall Procedure of DERN}
\label{alg:dern}
\begin{algorithmic}[1]
\Require $\mathcal{E} = \{E_1, \dots, E_N\}$, $\mathcal{C}$, $k$, $\alpha$
\Ensure $\mathcal{E}_r$

\State \textcolor{gray}{\textit{Stage 1: Expert Pruning}}
\For{$i = 1$ \textbf{to} $N$}
    \State $S_i \gets \mathbb{E}_{x_m \sim \mathcal{C}}\left[ \frac{\mathbb{I}[i \in \mathcal{A}(x_m)] \cdot g_i(x_m)}{\sum_{j \in \mathcal{A}(x_m)} g_j(x_m)} \right]$
\EndFor
\State $\mathcal{E}_r \gets \{ E_i \mid S_i \in \text{Top-}k(S_1, \dots, S_N) \}$
\State \textcolor{gray}{\textit{Stage 2: Decomposition, Recombination}}
\State $\mathcal{P} \gets \bigcup_{E_i \in \mathcal{E}_r} \text{Decompose}(E_i)$
\For{$\mathrm{seg}_i \in \mathcal{P}$}
    \State $E_r^* \gets \arg\max_{E_r \in \mathcal{E}_r} \text{sim}(\mathrm{seg}_i, E_r)$
    \If{$\text{sim}(\mathrm{seg}_i, E_r^*) > \alpha$}
        \State $E_r^* \gets E_r^* \cup \mathrm{seg}_i$
    \EndIf
\EndFor

\State \textcolor{gray}{\textit{Stage 3: Expert Reconstruction}}
\For{each $E_r \in \mathcal{E}_r$}
    \State $E_r' \gets \text{Cluster}(\mathcal{S}_r)$
\EndFor

\State \textbf{return} $\mathcal{E}_r$
\end{algorithmic}
\end{algorithm}

\subsection{Expert Reconstruction via Segment Clustering}
\label{sec:clustering}

After reassignment, each retained expert \( E_r \) is associated with a unified set of neuron segments:
\begin{equation}
\label{eq:Er_int_ext}
E_r \rightarrow \mathcal{S}_r = \mathcal{S}_r^{\text{int}} \cup \mathcal{S}_r^{\text{ext}},
\end{equation}
where \( \mathcal{S}_r^{\text{int}} \) contains the expert’s original segments, and \( \mathcal{S}_r^{\text{ext}} \) includes segments reassigned from pruned experts based on local structural similarity. To reduce redundancy while preserving diversity, we apply spherical weighted $k$-Means clustering to compress this combined segment set into a smaller, semantically coherent set of representative neurons.

We minimize a weighted cosine-based clustering objective over the segment set, using cluster centers \( \mathcal{C} = \{c_1, \dots, c_k\} \), where $k$ is the target hidden dimension of the reconstructed expert:
\begin{equation}
\label{eq:expert-cluster-objective}
\min_{\mathcal{C}} \sum_{j=1}^{k} \sum_{\mathrm{seg}_i \in \mathcal{S}_j} w_i \left( 1 - \frac{v(\mathrm{seg}_i)^\top c_j}{\|v(\mathrm{seg}_i)\| \|c_j\|} \right),
\end{equation}
where \( v(\mathrm{seg}_i) \) is the vectorized form of \( \mathrm{seg}_i \) in Eq.~\eqref{eq:vectorized-seg}, and \( w_i \) reflects the importance of \( \mathrm{seg}_i \), derived from its source expert's importance score.

To initialize cluster centers, we select the top-$k$ segments with the highest estimated activation bounds, measured by the maximum absolute values of their gate projection vectors \( w_{g,i} \). This leverages the gating mechanism’s implicit bias toward highly responsive neurons, promoting faster convergence and more coherent clusters.

To stabilize iteration and prevent segments with extremely large norms from dominating, we apply norm equalization before computing each cluster center. For each segment \( \mathrm{seg}_i \in S_j \), we scale its vector as:
\begin{align}
\hat{v}_i &= v(\mathrm{seg}_i) \cdot \frac{\bar{r}_j}{\|v(\mathrm{seg}_i)\|}, \nonumber\\
\bar{r}_j &= \frac{1}{|\mathcal{S}_j|} \sum_{\mathrm{seg}_k \in \mathcal{S}_j} \|v(\mathrm{seg}_k)\|.
\end{align}

Then, each cluster center \( c_j \) is computed via normalized weighted averaging:
\begin{align}
\label{eq:expert-cluster-center}
c_j &\leftarrow 
\frac{\sum_{\mathrm{seg}_i \in \mathcal{S}_j} \tilde{w}_i \hat{v}_i}
     {\left\| \sum_{\mathrm{seg}_i \in \mathcal{S}_j} \tilde{w}_i \hat{v}_i \right\|}, \nonumber\\
\tilde{w}_i &= \frac{w_i}{\sum_{\mathrm{seg}_k \in \mathcal{S}_j} w_k}.
\end{align}

Finally, the resulting cluster centers \( \{c_1, \dots, c_k\} \) form the compressed expert:
\begin{align}
\label{eq:reconstruct-expert-from-cluster-center}
E_r' \leftarrow \{c_1, c_2, \dots, c_k\},
\end{align}
offering a more compact and efficient representation while retaining the expert’s functional diversity and expressiveness.

\section{Experiments}

\begin{table*}[tb!]
\centering
\caption{Performance for Mixtral-8x7b-Instruct and Qwen2-57B-A14B-Instruct. Our method is highlighted in gray.}
\label{tab:mixtral_qwen}
\resizebox{\linewidth}{!}{%
\begin{tabular}{lccccccccccl}
\toprule
\textbf{Model} & \textbf{Method} & \textbf{PIQA} & \textbf{BoolQ} & \textbf{HellaS.} & \textbf{ARC-e} & \textbf{ARC-c} & \textbf{OBQA} & \textbf{OBQA-F} & \textbf{WinoG.} & \textbf{MMLU}  & \textbf{Avg.}\\
\midrule
\multicolumn{12}{c}{\textit{\textbf{Mixtral-8x7b-Instruct}}} \\
\midrule
\multirow{1}{*}{8*7B} 
& None & 82.75 & 78.69 & 80.68 & 92.24 & 87.46 & 83.20 & 89.40 & 64.48 & 69.01  &80.88 
\\
\midrule

\multirow{4}{*}{6*7B} 
& LLM‑Pruner & 75.63 & 77.49 & 69.53 & 85.71 & 76.27 & 71.80 & 89.00 & 57.30 & 59.30  &73.56 
\\
& M-SMoE & 68.66 & 80.24 & 57.40 & 85.36 & 70.17 & 70.80 & 86.80 & 55.17 & 52.44  &69.67 
\\
& NAEE & \textbf{82.59} & 77.22 & 74.71 & 88.01 & 80.68 & 77.00 & 87.40 & 63.61 & 64.00  &77.25 
\\
& \cellcolor[gray]{0.9}\textbf{DERN} & \cellcolor[gray]{0.9}81.72 & \cellcolor[gray]{0.9}\textbf{86.33} & \cellcolor[gray]{0.9}\textbf{78.48} & \cellcolor[gray]{0.9}\textbf{91.71} & \cellcolor[gray]{0.9}\textbf{82.03} & \cellcolor[gray]{0.9}\textbf{80.20} & \cellcolor[gray]{0.9}\textbf{90.20} & \cellcolor[gray]{0.9}\textbf{64.63} & \cellcolor[gray]{0.9}\textbf{65.10} & \cellcolor[gray]{0.9}\textbf{80.04} \\
\midrule

\multirow{4}{*}{4*7B} 
& LLM‑Pruner & 59.96 & 71.47 & 51.60 & 64.90 & 51.86 & 49.40 & 68.20 & 39.94 & 40.73  &55.34 
\\
& M-SMoE & 49.56 & 62.11 & 26.01 & 47.62 & 35.93 & 35.20 & 48.80 & 48.86 & 28.29  &42.49 
\\
& NAEE & 69.59 & 71.83 & \textbf{63.31} & 77.78 & 67.46 & 61.00 & 74.40 & 53.91 & 51.89  &65.69 
\\
& \cellcolor[gray]{0.9}\textbf{DERN} & \cellcolor[gray]{0.9}\textbf{70.08} & \cellcolor[gray]{0.9}\textbf{82.57} & \cellcolor[gray]{0.9}62.73 & \cellcolor[gray]{0.9}\textbf{83.42} & \cellcolor[gray]{0.9}\textbf{73.56} & \cellcolor[gray]{0.9}\textbf{71.80} & \cellcolor[gray]{0.9}\textbf{83.60} & \cellcolor[gray]{0.9}\textbf{54.46} & \cellcolor[gray]{0.9}\textbf{54.91} & \cellcolor[gray]{0.9}\textbf{70.79} \\
\midrule
\multicolumn{12}{c}{\textit{\textbf{Qwen2-57B-A14B-Instruct}}} \\
\midrule
\multirow{1}{*}{57.4B} 
& None & 84.00& 89.24& 87.08& 97.18& 91.19& 87.60& 93.60& 69.85& 75.54 &86.14 
\\
\midrule
\multirow{4}{*}{45.1B} 
& LLM‑Pruner & 76.93& 85.99& 82.38& 93.12& 86.10& 83.20& 91.80& 66.77& 70.64
 &81.88 
\\
& M-SMoE& 82.70& 87.09& 85.00& 94.53& 89.83& 83.40& 90.20& 64.80& 70.65
 &83.13 
\\
& NAEE& 80.09& 87.19& 85.05& 94.36& 88.14& 82.20& 92.00& 67.09& 72.09
 &83.13 
\\
& \cellcolor[gray]{0.9}\textbf{DERN} & \cellcolor[gray]{0.9}\textbf{84.60} & \cellcolor[gray]{0.9}\textbf{88.10} & \cellcolor[gray]{0.9}\textbf{86.02} & \cellcolor[gray]{0.9}\textbf{95.94} & \cellcolor[gray]{0.9}\textbf{90.17} & \cellcolor[gray]{0.9}\textbf{88.40} & \cellcolor[gray]{0.9}\textbf{93.60} & \cellcolor[gray]{0.9}\textbf{70.64} & \cellcolor[gray]{0.9}\textbf{74.06} & \cellcolor[gray]{0.9}\textbf{85.73} \\
\midrule
\multirow{4}{*}{33.3B} 
& LLM‑Pruner & 73.78& 81.25& 76.18& 76.19& 67.46& 67.00& 83.80& 60.77& 63.21
 &72.18 
\\
& M-SMoE& 70.89& 83.88& 73.76& 84.83& 69.83& 68.00& 82.20& 60.69& 58.33
 &72.49 
\\
& NAEE& 68.34& 79.24& 63.58& 86.24& 70.85& 64.80& 81.60& 62.12& 59.61
 &70.71 
\\
& \cellcolor[gray]{0.9}\textbf{DERN} & \cellcolor[gray]{0.9}\textbf{84.44} & \cellcolor[gray]{0.9}\textbf{87.03} & \cellcolor[gray]{0.9}\textbf{85.67} & \cellcolor[gray]{0.9}\textbf{92.77} & \cellcolor[gray]{0.9}\textbf{87.12} & \cellcolor[gray]{0.9}\textbf{85.60} & \cellcolor[gray]{0.9}\textbf{91.20} & \cellcolor[gray]{0.9}\textbf{67.01} & \cellcolor[gray]{0.9}\textbf{71.17} & \cellcolor[gray]{0.9}\textbf{83.56} \\
\bottomrule
\end{tabular}
}
\end{table*}

\subsection{Experimental Settings}

\paragraph{Model Settings.}
We evaluate our method on Mixtral-8$\times$7B-Instruct~\cite{jiang2024mixtral}, Qwen2-57B-A14B-Instruct~\cite{yang2024qwen2technicalreport}, and DeepSeek-MoE-16B-Chat~\cite{dai2024deepseekmoe}. For Mixtral (46.7B), we evaluate 6/8 (35.4B) and 4/8 (24.2B) expert configurations. For Qwen2 (57.4B), we use 48/64 (45.1B) and 32/64 (33.3B) expert configurations. For DeepSeek (16.4B), we use 56/64 (14.5B) and 48/64 (12.7B) configurations. These settings cover diverse architectures and compression ratios. All experiments were conducted on 2$\times$NVIDIA H100 GPUs.

\paragraph{Evaluation and Datasets.}
To support task-agnostic pruning, we calibrate pruning-related statistics using 128 sequences (each with 2048 tokens) randomly sampled from the C4 corpus~\cite{raffel2020exploring}, following Wanda~\cite{sun2023simple}.

We adopt the evaluation protocol of LLM-Pruner~\cite{ma2023llm}, conducting zero-shot evaluations on a suite of commonsense reasoning datasets: BoolQ~\cite{clark2019boolq}, PIQA~\cite{bisk2020piqa}, HellaSwag~\cite{zellers2019hellaswag}, ARC-e and ARC-c~\cite{clark2018think}, OpenbookQA~\cite{mihaylov2018can}, and WinoGrande~\cite{sakaguchi2021winogrande}. In addition, we evaluate multi-domain reasoning via few-shot prompting on MMLU~\cite{hendrycks2020measuring}.

All tasks are framed as generation-based evaluations, where the instruction-tuned model directly outputs the answer, and correctness is determined by template-based string matching. We follow OpenCompass~\cite{2023opencompass} for prompt formatting and matching criteria, and conduct all inference using the vLLM~\cite{kwon2023efficient}. For more details, see App.~\ref{appendix:evaluation}.

\paragraph{Baselines.}
We compare against representative structured SMoE pruning methods: NAEE (expert pruning)~\cite{lu2024not}, MC-SMoE (expert merging)~\cite{li2023merge}, and LLM-Pruner (general structured pruning)~\cite{ma2023llm}. For NAEE on Qwen2 and DeepSeek, we approximate expert selection by sampling 10k combinations per layer and choosing the one with minimal output deviation. For MC-SMoE, to ensure fairness under the retraining-free constraint, we disable intra-expert compression and denote it as M-SMoE. For LLM-Pruner, we evaluate on the taylor version, which is data-aware and gradient-based.

\subsection{Main Results}

\paragraph{Performance on Commonsense QA and Multidomain Benchmarks.}
\begin{table*}[tb]
\centering
\caption{Performance for DeepSeek-MoE-16b-Chat. Our method is highlighted in gray.}
\label{tab:ds}
\resizebox{\linewidth}{!}{%
    \begin{tabular}{lccccccccccl}
    \toprule
    \textbf{Model} & \textbf{Method} & \textbf{PIQA} & \textbf{BoolQ} & \textbf{HellaS.} & \textbf{ARC-e} & \textbf{ARC-c} & \textbf{OBQA} & \textbf{OBQA-F} & \textbf{WinoG.} & \textbf{MMLU}  & \textbf{Avg.}\\
    \midrule
    \multirow{1}{*}{16.4B} 
        & None & 67.57 & 75.75 & 54.00 & 70.55 & 49.83 & 47.40 & 68.80 & 56.99 & 48.32  &59.91 
\\
    \midrule
    \multirow{4}{*}{14.5B} 
        & LLM‑Pruner & 62.73 & 67.86 & 40.37 & 54.85 & 31.53 & 35.80 & 53.20 & 48.93 & 41.13  &48.49 
\\
        & M-SMoE & 60.28 & 61.19 & 43.90 & 50.62 & 37.63 & 37.80 & 52.80 & 51.22 & 39.13  &48.29 
\\
        & NAEE & \textbf{64.42} & 69.51 & 48.36 & 68.78 & 51.53 & 45.20 & \textbf{68.80} & 53.75 & 41.06  &56.82 
\\
        & \cellcolor[gray]{0.9}\textbf{DERN} & \cellcolor[gray]{0.9}61.75 & \cellcolor[gray]{0.9}\textbf{73.67} & \cellcolor[gray]{0.9}\textbf{52.75} & \cellcolor[gray]{0.9}\textbf{76.37} & \cellcolor[gray]{0.9}\textbf{55.93} & \cellcolor[gray]{0.9}\textbf{51.80} & \cellcolor[gray]{0.9}68.40 & \cellcolor[gray]{0.9}\textbf{53.91} & \cellcolor[gray]{0.9}\textbf{47.40} & \cellcolor[gray]{0.9}\textbf{60.22} \\

    \midrule
    \multirow{4}{*}{12.7B}
        & LLM‑Pruner & 54.68 & 26.36 & 26.59 & 34.74 & 21.02 & 26.00 & 36.20 & 47.59 & 34.05  &34.14 
\\
        & M-SMoE & 50.38 & 15.87 & 33.20 & 29.45 & 18.64 & 27.60 & 34.00 & 42.78 & 31.79  &31.52 
\\
        & NAEE & 56.37 & 52.14 & 34.35 & 55.73 & 36.27 & 39.80 & 59.40 & 51.38 & 34.60  &46.67 
\\
        & \cellcolor[gray]{0.9}\textbf{DERN} & \cellcolor[gray]{0.9}\textbf{60.83} & \cellcolor[gray]{0.9}\textbf{68.29} & \cellcolor[gray]{0.9}\textbf{44.47} & \cellcolor[gray]{0.9}\textbf{70.90} & \cellcolor[gray]{0.9}\textbf{48.14} & \cellcolor[gray]{0.9}\textbf{49.00} & \cellcolor[gray]{0.9}\textbf{60.20} & \cellcolor[gray]{0.9}\textbf{53.51} & \cellcolor[gray]{0.9}\textbf{44.17} & \cellcolor[gray]{0.9}\textbf{55.50} \\
    \bottomrule
    \end{tabular}
}
\vspace{-4mm}
\end{table*}

This section presents a comprehensive evaluation of our method against leading pruning baselines across three SMoE LLMs: Mixtral-8x7B-Instruct, Qwen2-57B-A14B-Instruct, and DeepSeek-MoE-16B-Chat.

Across all models and sparsity levels, DERN consistently outperforms existing techniques such as LLM-Pruner, M-SMoE, and NAEE, with particularly strong gains in commonsense reasoning and multidomain generalization. On Mixtral-8x7B-Instruct (Tab.~\ref{tab:mixtral_qwen}), DERN achieves leading performance on most Commonsense QA tasks, in some cases matching or surpassing the original dense model. With Qwen2-57B-A14B-Instruct, DERN maintains this advantage, outperforming all baselines on MMLU while preserving high accuracy on QA benchmarks, demonstrating strong resilience under compression.

\begin{figure}[tb]
    \centering
    \includegraphics[width=1\linewidth]{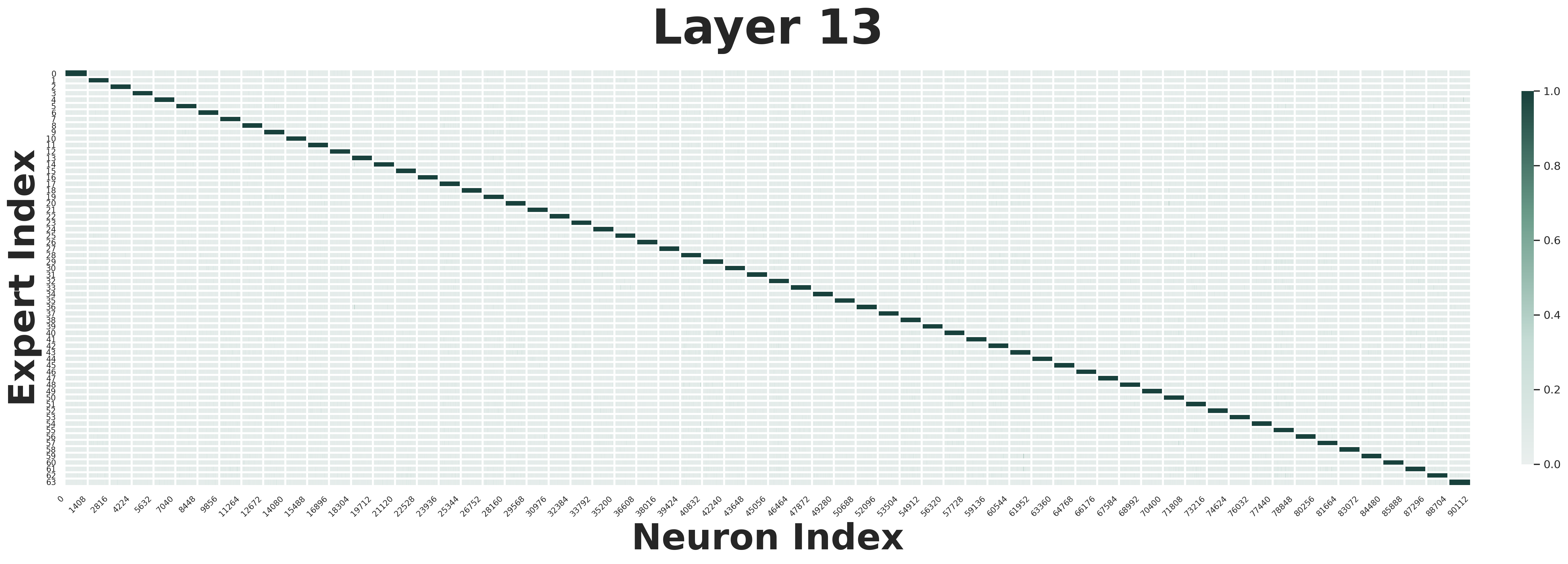}
    \caption{Neuron-level similarity heatmap among experts in Layer 13 of the DeepSeek-MoE-16b-Chat model. The predominantly light-colored off-diagonal regions highlight a high degree of independence and specialization among experts.}
    \label{fig:deepseek_sim_neuron_layer_13}
\vspace{-4mm}
\end{figure}

For DeepSeek-MoE-16B-Chat (Tab.~\ref{tab:ds}), DERN yields consistent gains at both 56 and 48 expert settings, outperforming baseline methods. However, as the pruning ratio increases, performance degradation becomes more pronounced across all methods. To better understand this behavior, we visualize expert similarity in Fig.~\ref{fig:deepseek_sim_neuron_layer_13}. The resulting heatmap reveals low inter-expert similarity, suggesting that DeepSeek’s experts are more independent and thus more vulnerable to pruning. This indicates that general-purpose expert pruning methods may require architectural adaptation to fully accommodate highly decoupled expert designs.

In summary, DERN delivers robust and scalable performance across diverse backbones and compression settings. Its superior knowledge preservation enables effective retention of both commonsense reasoning and cross-domain capabilities.

\paragraph{Inference Speedup and Memory Usage.}
Tab.~\ref{tab:memory_speed} reports inference efficiency on the vLLM before and after expert pruning. Reducing the number of experts from 8 to 6 and 4 yields notable improvements in both throughput and latency: token throughput increases by up to 38\%, while the time-to-first-token (TTFT) drops by over 900 ms. This acceleration primarily stems from improved kernel-level parallelism—fewer experts lead to more tokens routed to each active expert, resulting in larger and more GPU-efficient matrix multiplications. Meanwhile, memory usage is substantially reduced, enabling larger batch sizes and better hardware utilization. Notably, the 6-expert variant achieves nearly the same accuracy as the original model while offering 25\% lower memory consumption and 18\% higher throughput, demonstrating that DERN can effectively trade off redundancy for speed while maintaining competitive accuracy.

\begin{table}[tb]
\centering
\small  
\caption{Inference efficiency on Mixtral before and after expert pruning, evaluated on 1,024 C4 samples with 64-way concurrency using vLLM.}
\vspace{-2mm}
\label{tab:memory_speed}
\begin{tabular}{@{}lcccc@{}}
\toprule
\textbf{Model} & \textbf{Mem. (GB)} & \textbf{Tok/s} $\uparrow$ & \textbf{TTFT (ms)} $\downarrow$ & \textbf{Avg.} \\
\midrule
8$\times$7B & 88.7 & 7962.9 & 3349.5 & 80.88 \\
6$\times$7B & 66.0 & 9395.0 & 2850.0 & 80.04 \\
4$\times$7B & 45.0 & 10990.3 & 2443.8 & 70.60 \\
\bottomrule
\end{tabular}
\vspace{-4mm}
\end{table}

\begin{table*}[tb!]
\centering
\caption{
Ablation of different segment components used for similarity computation during segment reassignment. 
\textbf{DERN} uses only the up and down vectors; 
\textbf{triplet} includes gate, up, and down vectors; 
\textbf{w/o up} removes the up component; 
\textbf{w/o down} removes the down component.
}

\label{tab:segment-component-ablation}
\resizebox{\linewidth}{!}{%
    \begin{tabular}{lccccccccccl}
    \toprule
    \textbf{Model} & \textbf{Method} & \textbf{PIQA} & \textbf{BoolQ} & \textbf{HellaS.} & \textbf{ARC-e} & \textbf{ARC-c} & \textbf{OBQA} & \textbf{OBQA-F} & \textbf{WinoG.} & \textbf{MMLU}  & \textbf{Avg.}\\
    \midrule
    \multirow{4}{*}{Mixtral-4$\times$7B} 
        & \textbf{DERN} & 70.08& 82.57& 62.73& 83.42& 73.56& 71.8& 83.6& 54.46& 55.01 &70.80 \\
        & triplet& 70.57& 79.39& 59.43& 82.89& 72.88& 71.2& 84.2& 56.27& 55.11 &70.22 \\
        & w/o up& 71.98& 79.42& 58.9& 83.42& 72.54& 69.8& 83.2& 55.33& 54.74 &69.93 \\
        & w/o down& 70.57& 81.96& 59.49& 82.36& 71.86& 69.4& 85.2& 55.8& 55.51& 70.24 \\
    \bottomrule
    \end{tabular}
}
\vspace{-4mm}
\end{table*}

\subsection{Ablation Study}
\begin{figure}[tb]
    \centering
    \includegraphics[width=\linewidth]{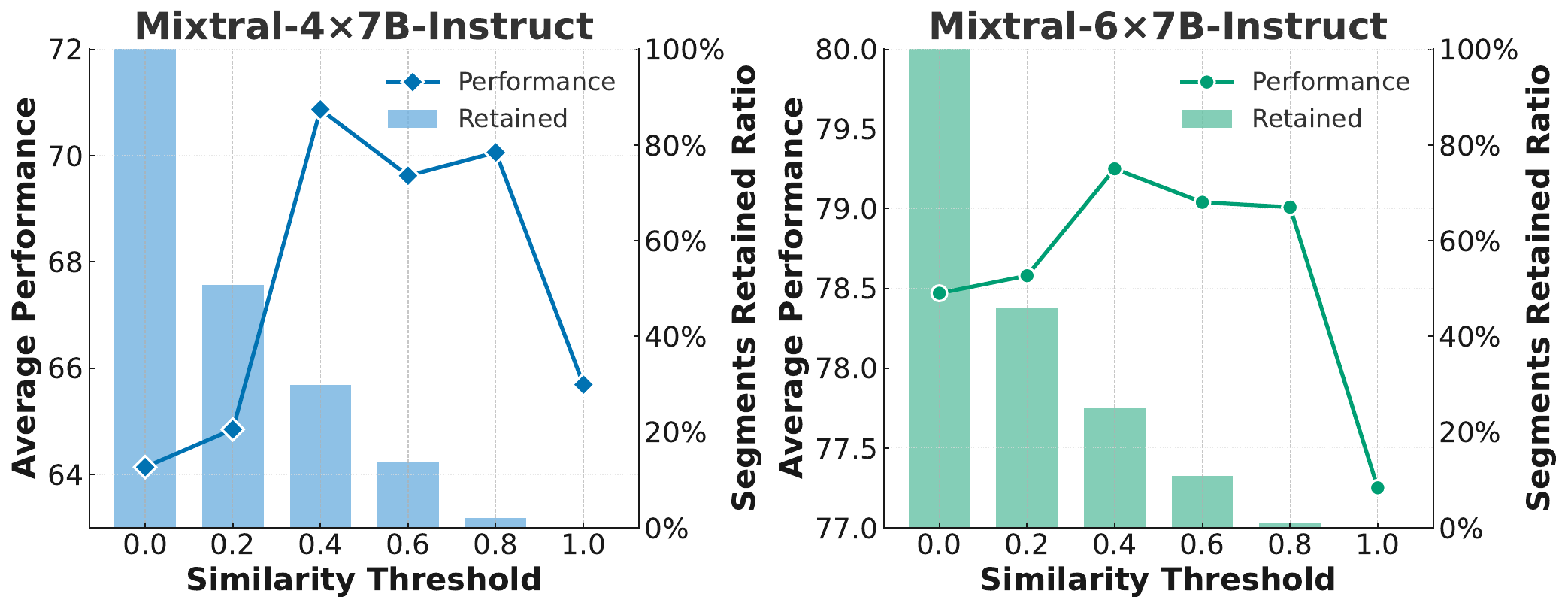}
    \caption{
    Effect of the similarity threshold $\alpha$ during segment reassignment. Line plots indicate average performance, and bars show the ratio of segments retained in last layer. Results are averaged over all benchmarks for Mixtral-4$\times$7B (left) and Mixtral-6$\times$7B (right).
    }
    \label{fig:sim-threshold-ablation}
\vspace{-4mm}
\end{figure}

\paragraph{Ablation on Segment Retention Ratio.}
Fig.~\ref{fig:sim-threshold-ablation} illustrates the effect of the similarity threshold $\alpha$ used in the second stage of DERN. Notably, $\alpha = 1$ disables merging entirely and serves as a no-merge baseline, while $\alpha = 0$ merges all pruned segments regardless of compatibility.

We observe a clear pattern across both Mixtral-4$\times$7B and Mixtral-6$\times$7B: performance improves as $\alpha$ increases from 0 to around 0.6, then drops sharply beyond that point. This reveals a fundamental trade-off: lower thresholds allow excessive reuse, introducing semantically mismatched segments; higher thresholds, while stricter, lead to underutilization. Best performance is achieved when only a moderate fraction of segments are reassigned, highlighting a “less is more” phenomenon where selective reuse outperforms indiscriminate merging. For more details, see App.~\ref{appendix:sim-albation}.

\paragraph{Ablation on Neuron Types Used in Similarity Estimation.}
Tab.~\ref{tab:segment-component-ablation} examines how neuron types affect similarity measurement in clustering. Each segment consists of a gate vector (activation), an up-projection vector (input transformation), and a down-projection vector (output transformation). Excluding the gate component slightly improves performance, suggesting dynamic activation signals may introduce noise. In contrast, removing either up or down leads to consistent drops, highlighting their complementary roles in capturing transformation behavior. The best results are achieved using only the up and down components—structural features that govern information flow. These findings support prioritizing weight geometry over activation dynamics in expert reconstruction.

\paragraph{Ablation on Clustering Initialization Strategy.}
Tab.~\ref{tab:init-ablation} compares three initialization methods for segment clustering: DERN (gate-based), which selects top-$k$ segments with the highest gate vector weights; Equidistant, which samples evenly across the similarity distribution; and Random choice. Gate-based initialization outperforms the others, showing that highly activated segments provide a more meaningful starting point for clustering. For more details, see App.~\ref{appendix:initialization}.

\begin{table}[tb]
\centering
\small
\caption{
Ablation on clustering initialization. 
\textbf{\textbf{DERN}} uses top-$k$ segments ranked by gate vector weights; 
\textbf{Random} samples seeds uniformly; 
\textbf{Equidistant} selects evenly from the similarity range.
}

\label{tab:init-ablation}
\begin{tabular}{lccc}
\toprule
\textbf{Strategy} & \textbf{CQA} & \textbf{MMLU} & \textbf{Average} \\
\midrule
\textbf{DERN}  & 72.78 & 55.01 & 63.89 \\
Random      & 58.93 & 38.29 & 48.61 \\
Equidistant & 72.08 & 55.03& 63.56 \\
\bottomrule
\end{tabular}
\end{table}

\paragraph{Ablation on Weighting Mechanism in Clustering.}
We evaluate whether weighting segments by their original expert importance improves cluster quality. As shown in Tab.~\ref{tab:weighting-ablation}, applying importance-based weighting during cluster center updates leads to a noticeable gain, suggesting that segments from more influential experts should be given higher priority in the reconstruction process.

\begin{table}[tb!]
\centering
\small
\caption{
Ablation on weighting mechanism in clustering. 
\textbf{DERN} applies segment importance during centroid updates; 
\textbf{w/o weighting} treats all segments equally.
}
\label{tab:weighting-ablation}
\begin{tabular}{lccc}
\toprule
\textbf{Strategy} & \textbf{CQA} & \textbf{MMLU} & \textbf{Average} \\
\midrule
\textbf{DERN}  & 72.78 & 55.01 & 63.89 \\
w/o weighting & 71.23 & 54.14 & 62.69 \\
\bottomrule
\end{tabular}
\vspace{-4mm}
\end{table}

\section{Conclusion}
We propose \textbf{DERN}, a task-agnostic and retraining-free framework for compressing SMoE LLMs. By treating experts as modular assemblies of functional segments, DERN performs decomposition and reconstruction based on neuron-level similarity. This enables compact expert representations that preserve model performance while reducing memory and latency. More broadly, DERN highlights a core insight: structural reorganization at the sub-expert level offers a robust and adaptable path to efficiency in SMoE.

\section*{Limitations}
While DERN performs well across various SMoE models, it measures segment similarity in parameter space using cosine distance. This may be insufficient for capturing functional alignment, especially in models with highly specialized experts. Notably, we observe that existing pruning and merging techniques perform suboptimally on DeepSeekMoE under high sparsity settings, where expert representations are more diverse and less interchangeable. This highlights the need for further investigation into pruning and recombination strategies tailored to highly specialized expert models.

\section*{Acknowledgments}
This work was supported in part by the National Science and Technology Major Project (2023ZD0120803), the National Natural Science Foundation of China (62472381), the Fundamental Research Funds for the Central Universities (226-2025-00055), the Fundamental Research Funds for the Zhejiang Provincial Universities (226-2024-00208), the "Pioneer" and "Leading Goose" R\&D Program of Zhejiang (2025C02032) and the Earth System Big Data Platform of the School of Earth Sciences, Zhejiang University.

\bibliography{custom}

\appendix
\clearpage
\section*{Appendix}

\section{Implementation Details}
\label{appendix:implementation}

\paragraph{Model Configuration.}
Our method is applied to three representative SMoE LLMs: Mixtral-8×7B-Instruct, Qwen2-57B-A14B-Instruct, and DeepSeek-MoE-16B-Chat. Each expert in these models is a gated MLP using the GLU architecture, parameterized by three projection matrices \( W_g \in \mathbb{R}^{h \times d} \), \( W_u \in \mathbb{R}^{h \times d} \), and \( W_d \in \mathbb{R}^{d \times h} \), where \( d \) and \( h \) denote the input and hidden dimensions respectively.

For Mixtral, we evaluate configurations with 8, 6, and 4 experts per MoE layer, using top-2 routing. Qwen2 and DeepSeek follow a similar sparsification scheme, with expert counts pruned from 64 to 48 or 32, and from 64 to 56 or 48 respectively. Unless otherwise noted, all models adopt top-$k$ routing with \( k = 2 \). A summary of expert-related configuration parameters is shown in Tab.~\ref{tab:model-config}.

\begin{table}[h]
\centering
\footnotesize
\caption{Comparison of expert configurations in Mixtral-8×7B-Instruct, Qwen2-57B-A14B-Instruct, and DeepSeek-MoE-16B-Chat.}
\begin{adjustbox}{width=\linewidth}
\begin{tabular}{lccc}
\toprule
\textbf{Property} & \textbf{Mixtral} & \textbf{Qwen2} & \textbf{DeepSeek} \\
\midrule
\texttt{hidden\_size}       & 4096 & 3584 & 2048 \\
\texttt{moe\_intermediate\_size}& 14336& 2560 & 1408 \\
\texttt{num\_hidden\_layers}& 32   & 28   & 28   \\
\texttt{num\_experts}       & 8    & 64   & 64   \\
\texttt{experts\_per\_token}& 2    & 8    & 6    \\
\texttt{shared\_experts}    & 0    & 1    & 2    \\
\texttt{activation}         & SiLU & SiLU & SiLU \\
\texttt{dtype}                   & bfloat16& bfloat16& bfloat16\\
\bottomrule
\end{tabular}
\end{adjustbox}
\label{tab:model-config}
\end{table}

\paragraph{Routing Statistics Collection.}
To evaluate expert importance, we support both parameter-based and data-driven strategies. In the data-driven setting, routing statistics are collected from a held-out calibration set consisting of 128 sequences sampled from the C4 corpus, each containing 2048 tokens. For each token, we record the top-$k$ experts selected by the gating mechanism along with their activation weights. These statistics are aggregated to reflect both the frequency and strength of each expert's participation, forming a routing-aware importance score as defined in Equation~\ref{eq:importance}. This calibration-aware pruning process enables adaptive selection of essential experts under realistic usage scenarios.

\paragraph{Segment Representation.}
Each expert is decomposed into a set of structurally independent segments, each corresponding to a triplet of projection vectors from the gate, up, and down components of the MLP block. These vectors are flattened and concatenated into fixed-dimensional representations, enabling efficient comparison across experts. To facilitate flexible merging, we define a similarity metric over segments based on weighted cosine distance, where the contribution of each component can be tuned. This design supports ablation studies over different segment configurations and allows fine-grained control over expert recombination.

\paragraph{Clustering Settings.}
Once segments are reassigned to retained experts based on local similarity, we apply spherical weighted $k$-means clustering~\cite{dhillon2001concept} to reduce redundancy and form compact expert representations. The number of output segments is determined proportionally to the original expert size, allowing for flexible compression. Cluster initialization prioritizes highly activated segments to promote stability and convergence. Segment representations are normalized before aggregation, and weighted updates are applied to emphasize contributions from more influential experts. This clustering process ensures both expressiveness and efficiency in the reconstructed model.

\paragraph{Hardware and Inference Setup.}
All experiments are performed on 2$\times$NVIDIA H100 GPUs using the vLLM serving framework~\cite{kwon2023efficient}. Inference is conducted in float16 precision with a batch size of 128. Inference efficiency is measured with 1,024 samples under 64-way concurrency. For evaluation, we follow the OpenCompass~\cite{2023opencompass} protocol, using generation-based decoding with prompt templates and string-matching for answer accuracy.

\section{Evaluation Details}
\label{appendix:evaluation}

\paragraph{Benchmarks.}
We evaluate our method on a diverse suite of benchmarks covering commonsense reasoning, scientific QA, and multi-domain knowledge understanding. Specifically, we include PIQA, BoolQ, HellaSwag, ARC-Easy, ARC-Challenge, OpenBookQA, and WinoGrande. For multi-domain evaluation, we adopt the MMLU benchmark, which contains 57 subjects ranging from STEM to humanities and professional domains.

\paragraph{Evaluation Setting.}
Commonsense QA tasks are conducted in the zero-shot setting, where no in-context examples are provided. For MMLU, we adopt a fixed 5-shot setting using a static in-context example retriever. Across all tasks, model predictions are evaluated via template-matching, where string-based rules extract the first valid option or capitalized answer token from the generated output.

\paragraph{Prompt and Evaluation Protocol.}
We use OpenCompass as the unified evaluation framework. Each dataset is paired with a custom prompt template designed for multiple-choice answering. Inference is conducted via autoregressive generation using a standard decoding configuration. Model outputs are post-processed by rule-based functions to extract answers in a structured and comparable format. Evaluation is performed using accuracy-based metrics, optionally supporting detailed per-category breakdowns. 

\tcbset{
  width=\linewidth, 
  halign=left,
  sharp corners,
  colback=gray!5,
  colframe=black!30,
  fonttitle=\bfseries,
  left=1mm, right=1mm, top=1mm, bottom=1mm
}

\begin{tcolorbox}[title=PIQA (Zero-shot Template)]
\ttfamily\footnotesize
\noindent
\{goal\} \\
A. \{sol1\} \\
B. \{sol2\} \\
Answer:
\end{tcolorbox}

\begin{tcolorbox}[title=BoolQ (Zero-shot Template)]
\ttfamily\footnotesize
\noindent
\{passage\} \\
Question: \{question\} \\
A. Yes \\
B. No \\
Answer:
\end{tcolorbox}

\begin{tcolorbox}[title=HellaSwag (Zero-shot Template)]
\ttfamily\footnotesize
\noindent
\{ctx\} \\
Question: Which ending makes the most sense? \\
A. \{A\} \\
B. \{B\} \\
C. \{C\} \\
D. \{D\} \\
Answer:
\end{tcolorbox}

\begin{tcolorbox}[title=ARC (Easy / Challenge) (Zero-shot Template)]
\ttfamily\footnotesize
\noindent
Question: \{question\} \\
A. \{textA\} \\
B. \{textB\} \\
C. \{textC\} \\
D. \{textD\} \\
Answer:
\end{tcolorbox}

\begin{tcolorbox}[title=OpenBookQA (Zero-shot Template)]
\ttfamily\footnotesize
\noindent
Question: \{question\_stem\} \\
A. \{A\} \\
B. \{B\} \\
C. \{C\} \\
D. \{D\} \\
Answer:
\end{tcolorbox}

\begin{tcolorbox}[title=OpenBookQA-Fact (Zero-shot Template)]
\ttfamily\footnotesize
\noindent
Given the fact: \{fact1\} \\
Question: \{question\_stem\} \\
A. \{A\} \\
B. \{B\} \\
C. \{C\} \\
D. \{D\} \\
Answer:
\end{tcolorbox}

\begin{tcolorbox}[title=WinoGrande (Zero-shot Template)]
\ttfamily\footnotesize
\noindent
Question: \{prompt\} \\
A. \{only\_option1\} \\
B. \{only\_option2\} \\
Answer:
\end{tcolorbox}

\begin{tcolorbox}[title=MMLU (5-shot Template)]
\ttfamily\footnotesize
\noindent
</E> \\
There is a single choice question about \{subject\}. Answer the question by replying A, B, C or D. \\

Question: \{example\_input\_1\} \\
A. \{A1\} \\
B. \{B1\} \\
C. \{C1\} \\
D. \{D1\} \\
Answer: \{label1\} \\[2pt]

\ldots \\[2pt]

Question: \{example\_input\_5\} \\
A. \{A5\} \\
B. \{B5\} \\
C. \{C5\} \\
D. \{D5\} \\
Answer: \{label5\} \\

</E> \\
There is a single choice question about \{subject\}. Answer the question by replying A, B, C or D. \\

Question: \{input\} \\
A. \{A\} \\
B. \{B\} \\
C. \{C\} \\
D. \{D\} \\
Answer:
\end{tcolorbox}

\paragraph{Dataset Configuration in OpenCompass.}

We conduct all evaluations using the standardized dataset configurations provided by the OpenCompass benchmark suite, which ensures consistency across data splits, prompt formats, and evaluation protocols. Specifically, we follow the officially released validation or test splits and adopt the corresponding input-output schema defined by OpenCompass for each task. This ensures compatibility with its prompt-based inference engine and allows for seamless integration with multiple large language models.

To maintain reproducibility and fairness, we preserve the default prompt templates and postprocessing routines associated with each dataset. These include input field extraction, candidate option formatting (where applicable), and answer string normalization during evaluation. Tables~\ref{tab:dataset-io} and~\ref{tab:dataset-eval} provide a detailed summary of the structural mappings and evaluation strategies employed for each benchmark.

The specific dataset variants used in our experiments follow the official OpenCompass configuration identifiers, which define the exact prompt structure, evaluation scripts, and scoring logic. The datasets and their corresponding OpenCompass IDs are listed below:

\begin{itemize}
    \item \textbf{PIQA}: \texttt{piqa\_gen\_1194eb}
    \item \textbf{BoolQ}: \texttt{SuperGLUE\_BoolQ\_gen\_883d50}
    \item \textbf{HellaSwag}: \texttt{hellaswag\_gen\_6faab5}
    \item \textbf{ARC-easy}: \texttt{ARC\_e\_gen\_1e0de5}
    \item \textbf{ARC-challenge}: \texttt{ARC\_c\_gen\_1e0de5}
    \item \textbf{OpenBookQA}: \texttt{obqa\_gen\_9069e4}
    \item \textbf{WinoGrande}: \texttt{winogrande\_gen\_458220}
    \item \textbf{MMLU}: \texttt{mmlu\_gen\_4d595a}
\end{itemize}

These configurations reflect the latest stable task definitions in OpenCompass and have been widely adopted in prior benchmarking studies. By adhering to these official variants without modification, our evaluations remain fully reproducible and aligned with best practices in LLM benchmarking.

\vspace{0.5em}
\begin{table}[tb]
\centering
\footnotesize
\caption{Input and output field configurations for each dataset.}
\begin{adjustbox}{width=\linewidth}
\begin{tabular}{lcc}
\toprule
\textbf{Dataset} & \textbf{Input Columns} & \textbf{Output Column} \\
\midrule
PIQA         & goal, sol1, sol2           & answer      \\
BoolQ        & passage, question          & label       \\
HellaSwag    & ctx, A, B, C, D            & label       \\
ARC-E / C    & question, textA--D         & answerKey   \\
OBQA         & question\_stem, A--D       & answerKey   \\
OBQA-F       & question\_stem, A--D, fact1& answerKey   \\
WinoGrande   & prompt, only\_option1/2    & answer      \\
MMLU         & input, A--D                & target      \\
\bottomrule
\end{tabular}
\end{adjustbox}
\label{tab:dataset-io}
\end{table}

\vspace{1em}

\begin{table}[h]
\centering
\footnotesize
\caption{Task types, evaluation strategies, and output postprocessing rules.}
\begin{adjustbox}{width=\linewidth}
\begin{tabular}{lccc}
\toprule
\textbf{Dataset} & \textbf{Task Type} & \textbf{Eval Type} & \textbf{Postproc} \\
\midrule
PIQA         & Commonsense QA        & Accuracy & A/B \\
BoolQ        & Boolean QA            & Accuracy & Yes/No \\
HellaSwag    & Sentence Completion   & Accuracy & A--D \\
ARC-E / C    & Science MCQ           & Accuracy & A--D \\
OBQA         & Open-book QA          & Accuracy & A--D \\
OBQA-F       & Fact-augmented QA     & Accuracy & A--D \\
WinoGrande   & Coreference Reasoning & Accuracy & A/B \\
MMLU         & Multidomain MCQ       & Accuracy & A--D \\
\bottomrule
\end{tabular}
\end{adjustbox}
\label{tab:dataset-eval}
\end{table}

\paragraph{Inference with vLLM.}
All evaluations are conducted using the vLLM serving framework, which enables high-throughput and memory-efficient inference via paged attention and continuous batching. In OpenCompass, we adopt two integration modes with vLLM:

\begin{itemize}
    \item \textbf{Direct backend mode}: By specifying \texttt{-a vllm} in the evaluation command, OpenCompass directly loads and executes models using vLLM as a backend engine. No additional serving is required.
    
    \item \textbf{Remote serving mode}: For scalable or distributed evaluation, vLLM is launched as a RESTful API server via \texttt{vllm serve}, and OpenCompass connects using the \texttt{OpenAISDK} interface. This enables model-as-a-service deployment and decouples evaluation from inference hosting.
\end{itemize}

Each task is evaluated with a maximum sequence length of 256 tokens, batch size of 128, and deterministic decoding (\texttt{temperature = 0.0001}). Additional runtime parameters (e.g., tensor parallelism, GPU memory usage) are configured via model-level arguments. The following illustrates a typical vLLM server launch and OpenCompass API configuration:

\vspace{0.5em}
\noindent\textbf{Launching vLLM server:}
\begin{lstlisting}
vllm serve mistralai/Mixtral-8x7B-Instruct --host 0.0.0.0 --port 8000
\end{lstlisting}

\noindent\textbf{OpenCompass model config:}
\begin{lstlisting}[language=Python]
dict(
    abbr='mixtral-vllm-api',
    type=OpenAISDK,
    openai_api_base='http://localhost:8000/v1',
    path='mistralai/Mixtral-8x7B-Instruct',
    tokenizer_path='mistralai/Mixtral-8x7B-Instruct',
)
\end{lstlisting}

All evaluations are fully automated and seamlessly integrated with the pruning pipeline. Once a model is compressed, its checkpoint is registered into the evaluation system and executed using vLLM without manual intervention. This ensures consistency across settings and enables scalable assessment of compression-quality trade-offs.

\section{Further Experimental Analyses}
\label{appendix:further-analysis}

This section provides additional results and ablation studies that complement the main findings. It includes full benchmark breakdowns, detailed analysis of design choices, and supporting visualizations.

\subsection{More Visualization of Neuron-level Similarity}
\label{appendix:neuron-sim}

\begin{figure*}[ht]
    \centering
    \includegraphics[width=\linewidth]{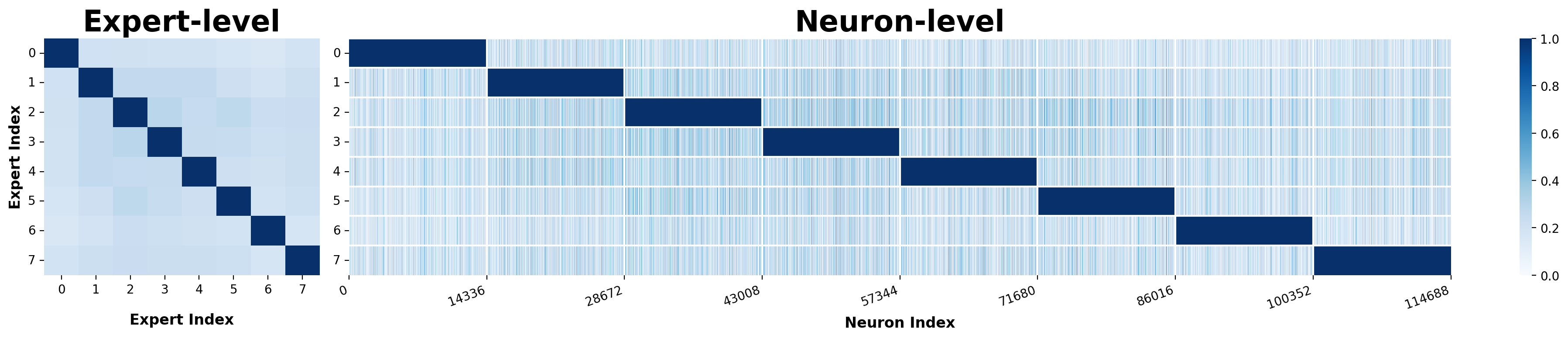}\vspace{-1em}
    
    \includegraphics[width=\linewidth]{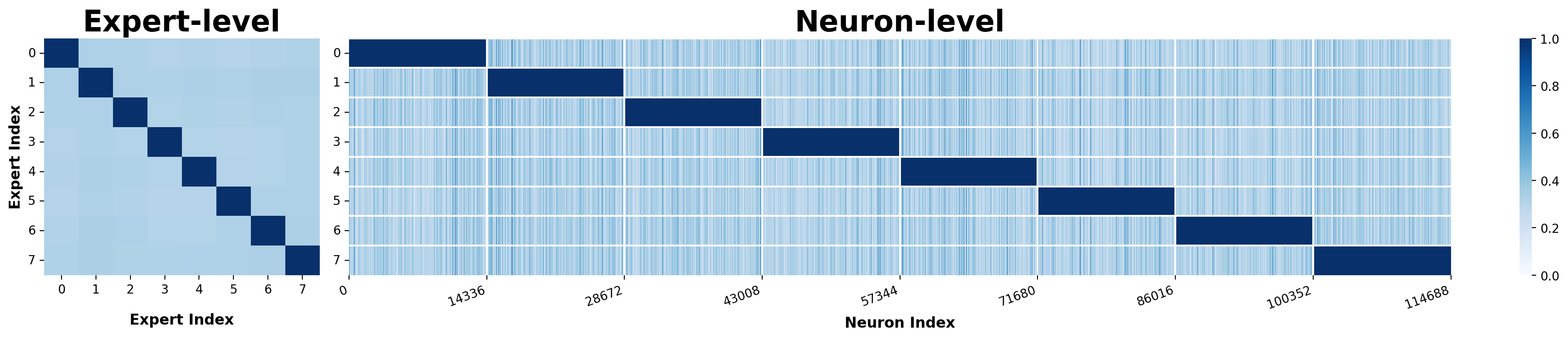}\vspace{-1em}
    
    \includegraphics[width=\linewidth]{fig/layer15_combined_similarity.png}\vspace{-1em}
    
    \includegraphics[width=\linewidth]{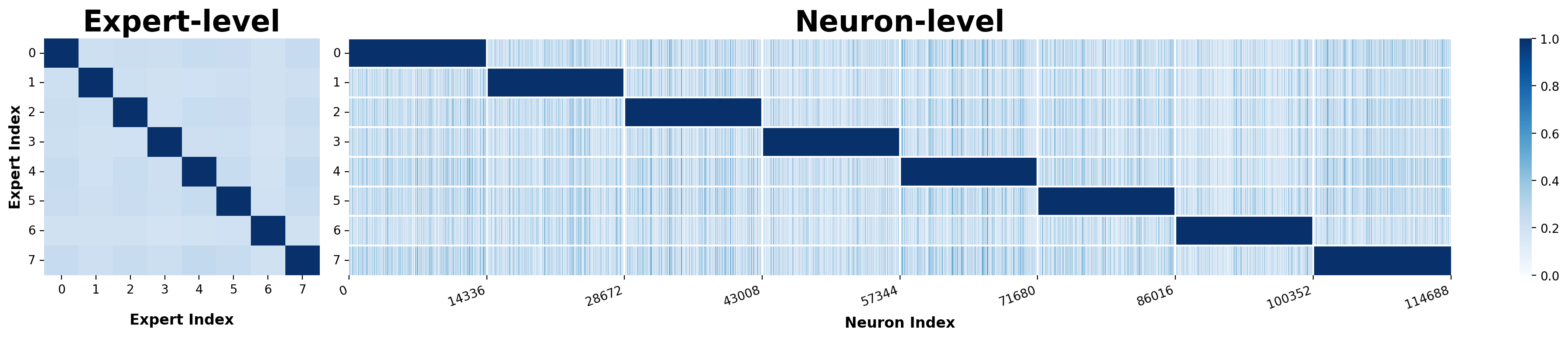}\vspace{-1em}
    
    \includegraphics[width=\linewidth]{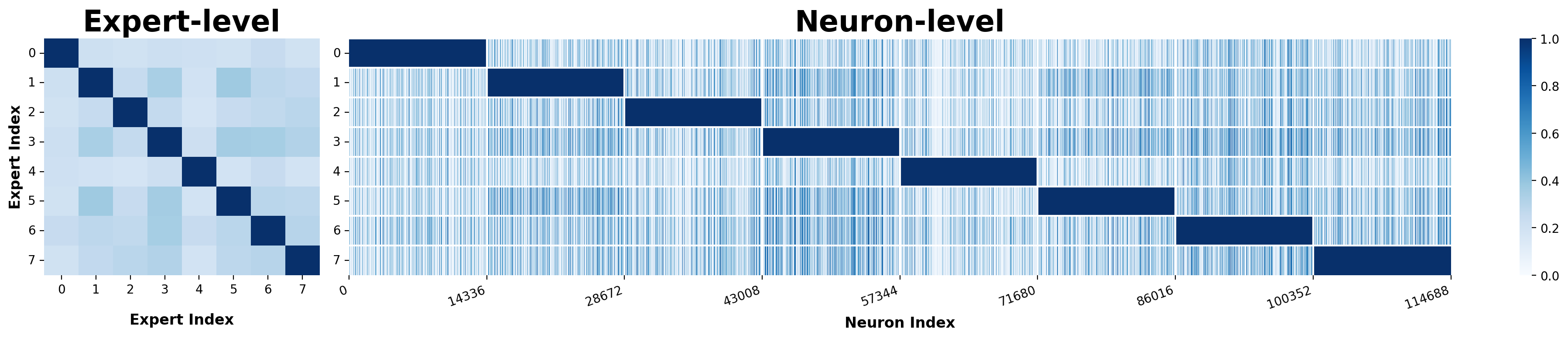}
    \caption{Neuron-level and expert-level similarity heatmaps across Layers 0, 7, 15, 23, and 31. Left: expert similarity; Right: neuron alignment.}
    \label{fig:all-layer-similarity}
\end{figure*}

To better understand the structural differences across experts in different layers, we visualize both expert-level and neuron-level similarity for several representative layers of the Mixtral-8×7B-Instruct model. As shown in Figures~\ref{fig:all-layer-similarity}, the left subplots represent cosine similarity between expert weight matrices, while the right subplots reflect neuron-level alignment, where each row shows the maximal similarity between neurons in a source expert and neurons in a target expert.

We observe several key trends:

\begin{itemize}
    \item In early layers (e.g., Layer 0), experts are highly distinct, as evidenced by strong diagonal blocks in expert-level similarity and sparse off-diagonal interactions at the neuron level.
    \item As depth increases (e.g., Layer 7 and 15), some inter-expert similarity begins to emerge. Neuron-level heatmaps reveal increasing degrees of overlap, suggesting redundancy or convergence in learned representations.
    \item In deeper layers (e.g., Layer 23 and 31), experts exhibit diffuse structural boundaries and higher degrees of entanglement at the neuron level, complicating naive expert merging.
\end{itemize}

These observations support our core motivation: expert-level similarity alone is insufficient for reliable merging, and neuron-level structural alignment is essential for preserving functionality. DERN’s segment-based recombination mechanism addresses this by explicitly matching and merging structurally compatible components.


\subsection{Full Results of Similarity Threshold $\alpha$ Ablation}
\label{appendix:sim-albation}
Table~\ref{tab:full-results-sim-alpha} provides the complete results of our ablation study on the similarity threshold $\alpha$, which controls segment reassignment during expert merging. We report results on Mixtral-8$\times$7B-Instruct under two configurations: 6-expert (35.4B) and 4-expert (24.2B).

We observe a consistent pattern across both settings: performance improves as $\alpha$ increases from 0 to around 0.4--0.6, but then sharply drops as $\alpha$ approaches 1. This trend reveals a key trade-off: lower thresholds enable more segment reuse, but risk semantic mismatches due to indiscriminate merging; higher thresholds impose stricter filtering, but lead to underutilization of transferable structure. Notably, the best performance is achieved when only a moderate fraction of segments are reused, confirming that \textit{selective reuse is superior to exhaustive merging}.

For the 4-expert setting, accuracy peaks at $\alpha=0.4$ (70.87 average), significantly outperforming the baseline ($64.14$ at $\alpha=0$), while reducing the active segment ratio to approximately 30\%. Similarly, in the 6-expert setting, $\alpha=0.4$ achieves the best trade-off between quality and compression, improving average performance from $78.47$ (baseline) to $79.25$. These results highlight a ``less is more'' phenomenon: \textit{reusing only the most structurally compatible segments suffices to reconstruct performant and efficient experts}.

This behavior aligns with our broader insight: DERN benefits from structure-aware modularity. By leveraging local segment similarity, the method avoids the pitfalls of whole-expert averaging while effectively capturing reusable computation patterns. As a design recommendation, we find that $\alpha \in [0.4, 0.6]$ offers a robust operational range across SMoE backbones.

\begin{table*}[tb!]
\centering
\small
\caption{Full Results of Similarity Threshold $\alpha$ Ablation on Mixtral-4$\times$7B-Instruct and Mixtral-6$\times$7B-Instruct.}
\label{tab:full-results-sim-alpha}
\begin{adjustbox}{max width=\textwidth}
\begin{tabular}{@{}llcccccccccc@{}}
\toprule
Setting & piqa & BoolQ & hellaswag & ARC-e & ARC-c & openbookqa & openbookqa\_fact & winogrande & mmlu & Avg. & ratio \\
\midrule
\multicolumn{12}{c}{\textbf{Mixtral-4$\times$7B-Instruct}} \\
\midrule
sim\_0   & 71.38 & 53.67 & 63.82 & 80.95 & 64.07 & 65   & 78.8 & 50.12 & 49.49 & 64.14 & 100\% \\
sim\_0.2 & 71.82 & 52.17 & 64.31 & 80.78 & 64.41 & 66.2 & 81.8 & 50.91 & 51.22 & 64.85 & 50.77\% \\
sim\_0.4 & 73.88 & 81.01 & 59.86 & 85.19 & 72.88 & 69.4 & 85   & 55.88 & 54.74 & 70.87 & 29.79\% \\
sim\_0.6 & 71.00 & 79.69 & 58.69 & 83.42 & 71.86 & 69   & 84.6 & 53.20 & 55.16 & 69.62 & 13.70\% \\
sim\_0.8 & 71.16 & 80.03 & 60.07 & 83.42 & 72.88 & 68.6 & 84.4 & 55.17 & 54.83 & 70.06 & 2.08\% \\
sim\_1.0 & 69.59 & 71.83 & 63.31 & 77.78 & 67.46 & 61   & 74.4 & 53.91 & 51.89 & 65.69 & 0\% \\
\midrule
\multicolumn{12}{c}{\textbf{Mixtral-6$\times$7B-Instruct}} \\
\midrule
sim\_0   & 81.77 & 84.98 & 77.63 & 90.48 & 80.00 & 77.8 & 89.0 & 61.88 & 62.68 & 78.47 & 100\% \\
sim\_0.2 & 78.94 & 85.26 & 77.61 & 90.48 & 80.34 & 79.8 & 89.2 & 62.51 & 63.12 & 78.58 & 45.99\% \\
sim\_0.4 & 80.20 & 86.76 & 78.31 & 91.18 & 81.36 & 79.4 & 89.2 & 62.27 & 64.53 & 79.25 & 25.15\% \\
sim\_0.6 & 79.98 & 85.32 & 78.08 & 91.89 & 82.71 & 78.2 & 88.4 & 62.27 & 64.50 & 79.04 & 10.79\% \\
sim\_0.8 & 80.69 & 84.77 & 78.00 & 91.18 & 82.37 & 78.2 & 88.6 & 62.67 & 64.60 & 79.01 & 1.10\% \\
sim\_1.0 & 82.59 & 77.22 & 74.71 & 88.01 & 80.68 & 77.0 & 87.4 & 63.61 & 64.00 & 77.25 & 0\% \\
\bottomrule
\end{tabular}
\end{adjustbox}
\end{table*}

\subsection{Initialization Strategy}
\label{appendix:initialization}
\begin{figure*}[htbp]
    \centering
    \includegraphics[width=0.8\linewidth]{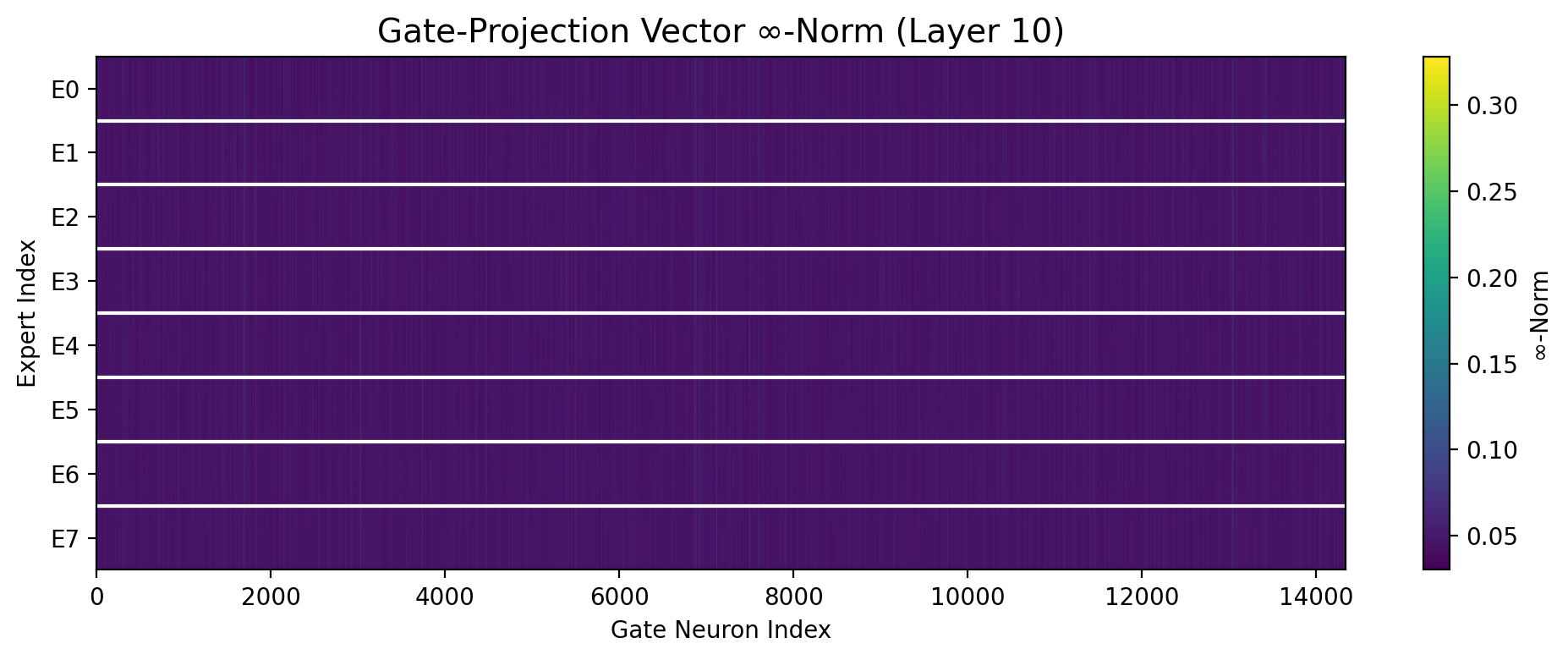}
    \vspace{-1mm}
    
    \includegraphics[width=0.8\linewidth]{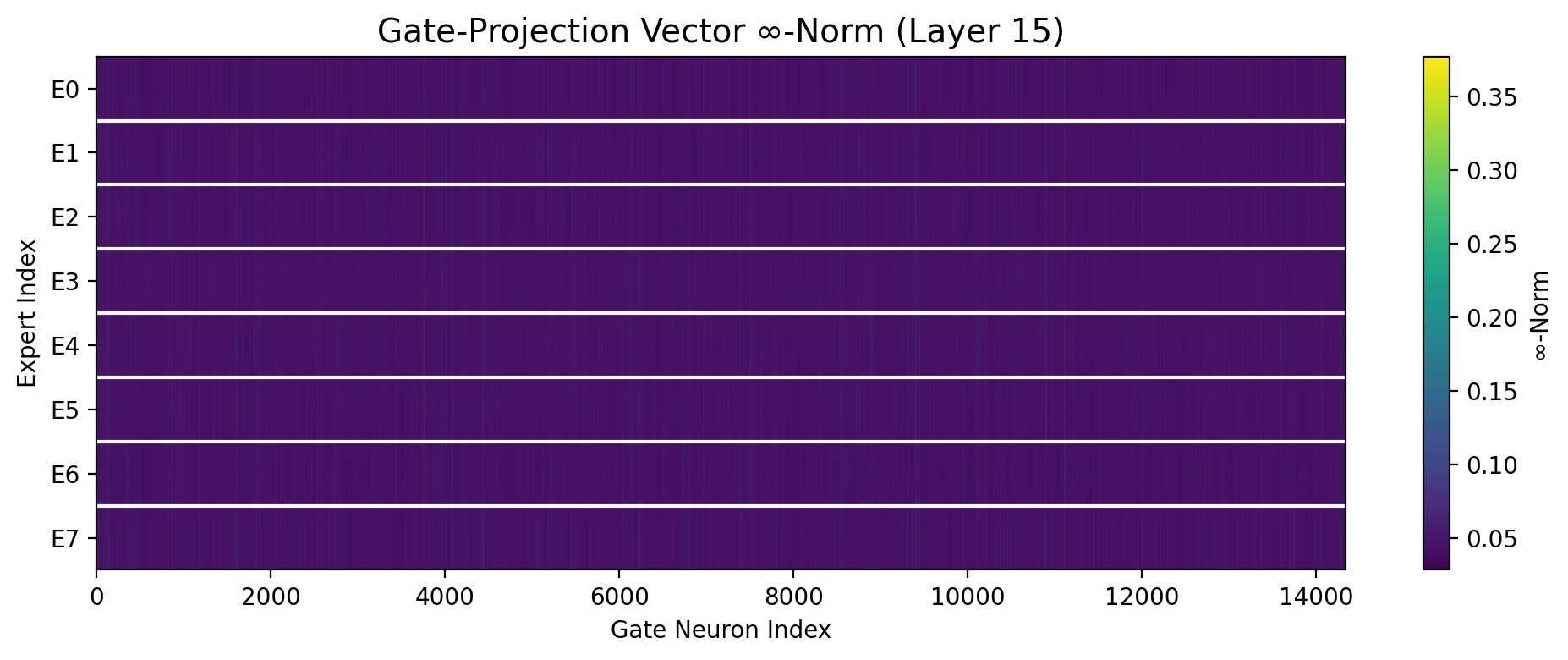}
    \vspace{-1mm}
    
    \includegraphics[width=0.8\linewidth]{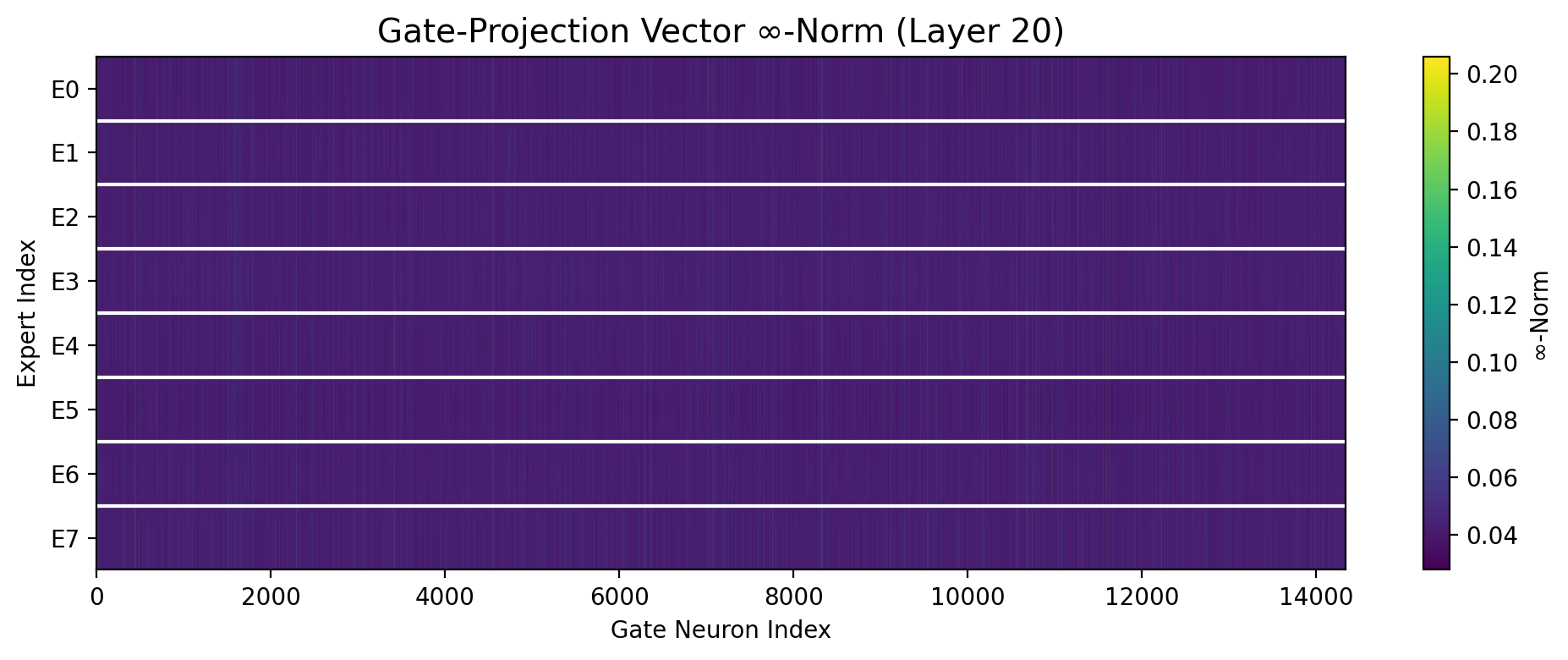}
    \vspace{-1mm}
    
    \includegraphics[width=0.8\linewidth]{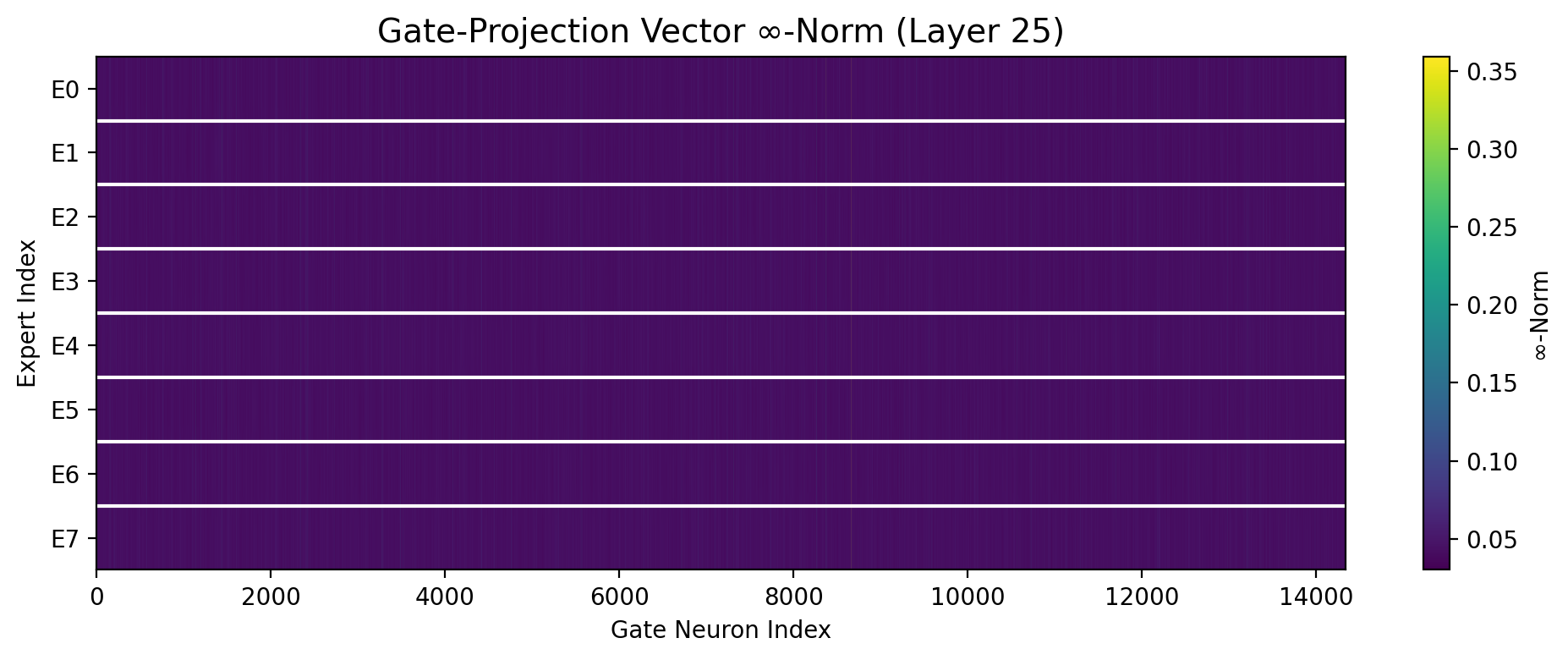}
    \vspace{-1mm}
    
    \caption{$\ell_\infty$ norm of row vectors $W_g$ across Layers 10, 15, 20, and 25.}
    \label{fig:inf-norm-multilayer}
\end{figure*}

\begin{table*}[tb!]
\centering
\small
\caption{Full Results on Ablation of different clustering initialization strategies on Mixtral-4$\times$7B-Instruct.}
\label{tab:clustering-init}
\begin{adjustbox}{max width=\textwidth}
\begin{tabular}{@{}lcccccccccc@{}}
\toprule
Strategy & PIQA & BoolQ & HellaSwag & ARC-e & ARC-c & OpenBookQA & OBQA-Fact & WinoGrande & MMLU & Avg. \\
\midrule
Gate-based   & 70.08 & 82.57 & 62.73 & 83.42 & 73.56 & 71.80 & 83.60 & 54.46 & 55.01 & 70.80 \\
Random       & 60.01 & 54.28 & 46.30 & 73.02 & 58.31 & 55.00 & 75.00 & 49.49 & 38.29 & 57.74 \\
Equidistant  & 71.38 & 81.53 & 61.56 & 82.72 & 73.22 & 69.00 & 83.20 & 54.06 & 55.03 & 70.30 \\
\bottomrule
\end{tabular}
\end{adjustbox}
\end{table*}

To support the segment clustering process in our expert reconstruction framework, we visualize the $\ell_\infty$ norm of each \textbf{row vector} in the gate projection matrix $W_g$ across four different Transformer layers. Each row vector in $W_g$ corresponds to the gating component of a segment and determines its activation magnitude via the term $\sigma(w_{g,i}^\top x)$, as formulated in Eq.~(3) of the main paper. This activation controls the contribution of the segment to the final expert output, modulating the up-projection vector.

The $\ell_\infty$ norm of each row provides a conservative upper bound on that segment’s gating strength over all possible inputs. Segments with higher $\ell_\infty$ norms are more likely to be activated with greater magnitude and thus play a more prominent role in computation. 

As illustrated in Figure~\ref{fig:inf-norm-multilayer}, different layers exhibit distinct distributions of segment-wise activation potential. Some segments are consistently dominant, while others have negligible norms, indicating a disparity in their functional contribution. We leverage this observation to guide the initialization of our clustering algorithm by selecting high-norm segments as cluster centroids. This ensures that structurally and functionally salient components are retained during expert merging, which is crucial for maintaining model performance after compression.

To further validate this design choice, we present the full results of different clustering initialization strategies in Table~\ref{tab:clustering-init}. Among the three evaluated strategies—gate-based, random, and equidistant—the gate-based method consistently outperforms others across tasks. This supports the intuition that segment saliency, as reflected by gating magnitude, serves as an effective heuristic for initialization during expert reconstruction.


\section{Link to Models and Datasets}
\label{sec:link}
All models and datasets used in this work are publicly available and sourced from established repositories to ensure reproducibility and transparency. We rely on widely adopted pretrained SMoE language models including Mixtral-8$\times$7B-Instruct, Qwen2-57B-A14B-Instruct, and DeepSeek-MoE-16B-Chat, all hosted on Hugging Face. These models represent a diverse set of sparse expert architectures and serve as representative backbones for evaluating the generality of our proposed method.

For dataset selection, we follow common benchmarks used in LLM pruning and evaluation literature. Specifically, we include the C4 corpus for calibration, and a suite of commonsense and multi-domain reasoning datasets such as BoolQ, PIQA, HellaSwag, ARC, OpenBookQA, WinoGrande, and MMLU. All datasets are retrieved through the Hugging Face Datasets Hub, ensuring consistent preprocessing and accessibility.

A full list of links to model checkpoints and dataset resources is provided below for reference.

\subsection{Models}
\begin{description}[style=unboxed, leftmargin=0pt, itemsep=2pt]
  \item[Mixtral-8$\times$7B-Instruct] 
  \url{https://huggingface.co/mistralai/Mixtral-8x7B-Instruct-v0.1}
  
  \item[Qwen2-57B-A14B-Instruct] 
  \url{https://huggingface.co/Qwen/Qwen2-57B-A14B-Instruct}
  
  \item[DeepSeek-MoE-16B-Chat] 
  \url{https://huggingface.co/deepseek-ai/deepseek-moe-16b-chat}
\end{description}

\subsection{Datasets}
\begin{description}[style=unboxed, leftmargin=0pt, itemsep=2pt]
  \item[C4] 
  \url{https://huggingface.co/datasets/allenai/c4}
  
  \item[BoolQ] 
  \url{https://huggingface.co/datasets/google/boolq}
  
  \item[PIQA] 
  \url{https://huggingface.co/datasets/ybisk/piqa}
  
  \item[HellaSwag] 
  \url{https://huggingface.co/datasets/Rowan/hellaswag}
  
  \item[ARC] 
  \url{https://huggingface.co/datasets/allenai/ai2_arc}
  
  \item[OpenBookQA] 
  \url{https://huggingface.co/datasets/allenai/openbookqa}
  
  \item[WinoGrande] 
  \url{https://huggingface.co/datasets/allenai/winogrande}
  
  \item[MMLU] 
  \url{https://huggingface.co/datasets/cais/mmlu}
\end{description}

\end{document}